\documentclass[10pt,twocolumn,letterpaper]{article}

\usepackage{wacv}              % To produce the CAMERA-READY version
\usepackage{graphicx}
\usepackage{amsmath}
\usepackage{amssymb}
\usepackage{booktabs}

\usepackage[accsupp]{axessibility}  % Improves PDF readability for those with disabilities.

\usepackage{url}            % simple URL typesetting
\usepackage{amsfonts}       % blackboard math symbols
\usepackage{nicefrac}       % compact symbols for 1/2, etc.
\usepackage{microtype}      % microtypography
\usepackage{xcolor}         % colors

\newcommand{\sj}[1]{{\textcolor{purple}{\bf \em [sj: #1]}}}

\usepackage[pagebackref,breaklinks,colorlinks]{hyperref}

\usepackage[capitalize]{cleveref}
\crefname{section}{Sec.}{Secs.}
\Crefname{section}{Section}{Sections}
\Crefname{table}{Table}{Tables}
\crefname{table}{Tab.}{Tabs.}

\def\wacvPaperID{328} % *** Enter the WACV Paper ID here
\def\confName{WACV}
\def\confYear{2024}

\begin{document}

\title{Edge Inference with Fully Differentiable Quantized Mixed Precision Neural Networks}

\author{
Clemens JS Schaefer\thanks{Work partly conducted while interning at Google LLC}$^{\;\;\dagger}$, Siddharth Joshi$^{\dagger}$, Shan Li$^{\ddagger}$ and Raul Blazquez$^{\ddagger}$\\
    $^{\dagger}$ University of Notre Dame, Notre Dame, IN, USA \\
    $^{\ddagger}$ Google LLC, Mountain View, CA, USA \\
    \textit{\{cschaef6, sjoshi2\}@nd.edu, \{lishanok, rblazquez\}@google.com}
% First Author\\
% \and
}
\maketitle

%%%%%%%%% ABSTRACT
\begin{abstract}
   The large computing and memory cost of deep neural networks (DNNs) often precludes their use in resource-constrained devices. Quantizing the parameters and operations to lower bit-precision offers substantial memory and energy savings for neural network inference, facilitating the use of DNNs on edge computing platforms. Recent efforts at quantizing DNNs have employed a range of techniques encompassing progressive quantization, step-size adaptation, and gradient scaling. This paper proposes a new quantization approach for mixed precision convolutional neural networks (CNNs) targeting edge-computing. Our method establishes a new Pareto frontier in model accuracy and memory footprint demonstrating a range of pre-trained quantized models, delivering best-in-class accuracy below 4.3~MB of weights and activations without modifying the model architecture. Our main contributions are: (i) a method for tensor-sliced learned precision with a hardware-aware cost function for heterogeneous differentiable quantization, (ii) targeted gradient modification for weights and activations to mitigate quantization errors, and (iii) a multi-phase learning schedule to address instability in learning arising from updates to the learned  quantizer and model parameters. We demonstrate the effectiveness of our techniques on the ImageNet dataset across a range of models including EfficientNet-Lite0 (e.g., 4.14~MB of weights and activations at 67.66\% accuracy) and MobileNetV2 (e.g., 3.51~MB weights and activations at 65.39\% accuracy).
\end{abstract}

%%%%%%%%% BODY TEXT

\section{Introduction}

% natural language processing (\cite{takase2021lessons}), recommendation systems (\cite{kim2019enhancing}) and even the protein folding problem (\cite{tunyasuvunakool2021highly})

Deep neural networks (DNNs) demonstrate remarkable performance at computer vision tasks,  notably being the defacto standard methods employed for large scale image recognition (\cite{zhai2021scaling, chen2021vision}). However, modern deep learning models require substantial compute and memory resources. This presents a challenge in deploying DNNs on resource constrained edge hardware.

Techniques for developing edge-deployable DNNs include the design of hardware-friendly DNN models, the development of low-power/latency hardware, model pruning, and quantizing DNNs to operate at lower precision\cite{tan2019efficientnet,sandler2018mobilenetv2, wan2021edge}. Since low-precision operations can simultaneously lower the memory footprint, increase throughput, and lower the latency for DNN inference, DNN quantization has become increasingly important~\cite{jacob2018quantization}. In particular, with recent DNN accelerators~\cite{yazdanbakhsh2021evaluation, jouppi2017datacenter} and graphics processing units (GPUs)~\cite{shoeybi2019megatron} offering support for mixed-precision computing, these benefits can be realized on existing hardware~\cite{yao2021hawq}. Furthermore, the compilers and hardware community are actively researching how to extend the support for multiple low-bit width mixed precision operations~\cite{garofalo2022darkside, risso2023precision, rutishauser2023free,molendijk2022braintta,mo2022motuner}. With additional compiler support for mixed precision quantization, such quantization could be leveraged for DNNs deployed on field programmable gate arrays (FPGAs). However, existing quantized DNN models do not fully leverage such hardware capabilities. In particular, most model quantization approaches focus on weight quantization, ignoring the high energy and latency costs of moving and storing activations. Benchmarking indicates that in intermediate layers, these costs dominate in accelerators~\cite{chen2019eyeriss}.

Previous research~\cite{yao2021hawq, uhlich2019mixed, wang2019haq} has shown promising results with heterogeneous quantization, allocating memory resources per layer. This form of quantization can facilitate a wider range of trade offs between networks size and accuracy. Building on this insight, we focus this paper on developing compact DNN models extremely low memory footprints. We report best-in-class results for models with a total memory footprint below 4.3~MB.

%  size far beyond what has previously been possible, e.g. most current state-of-the-art multi bit-width networks can only be compressed down to sizes around 

In this paper we: (i) present a hardware-aware mixed precision differentiable quantization formulation which includes per-tensor learned precision for activations and fine-grained per-channel quantization for weights, (ii) propose a novel gradient modification scheme which entails modifying weights and activation gradients differently and introduce arctanh based gradient scaling together with comprehensive evaluations against other gradient scaling techniques, and (iii) introduce a multi-phase learning-rate schedule to address instability in learning arising from updates to the learned quantizers and perform extensive comparisons of this schedule with alternatives. We show the effectiveness of our methods on the ImageNet dataset using the EfficientNet-Lite0, MobileNetV2, wide SqueezeNext, and ResNet18 model architectures. We demonstrate state-of-the-art accuracy for multi bit-width models ranging from 2.89--4.3 MB total memory footprint. Across various model architectures, our quantization scheme forms the Pareto optimal frontier for model accuracy vs. model size.

% including homogeneous pre-training, penalty scheduling, alternating phases of quantizer parameters updated and final model parameter fine-tuning
%In this paper we will analyze the effect of gradient scaling on quantization aware training of heterogeneous neural networks. Our main contributions are: i) a comprehensive training recipe for mixed precisions neural networks, ii) a thorough analysis of gradient scaling for quantized NN, iii) a presentation of a full state-of-the-art efficient frontier for full network size (sum of weights and sum of activations) and performance on the ImageNet dataset given various modern NN architectures.

% which in some cases even helps to prevent overfitting.  

% Hardware and custom accerators energy efficiency -> mixed precision DNNs optimizing them for energy consumption.

\section{Background}\label{sec:back}

\begin{figure*}
  \centering
  \includegraphics[width=.7\textwidth]{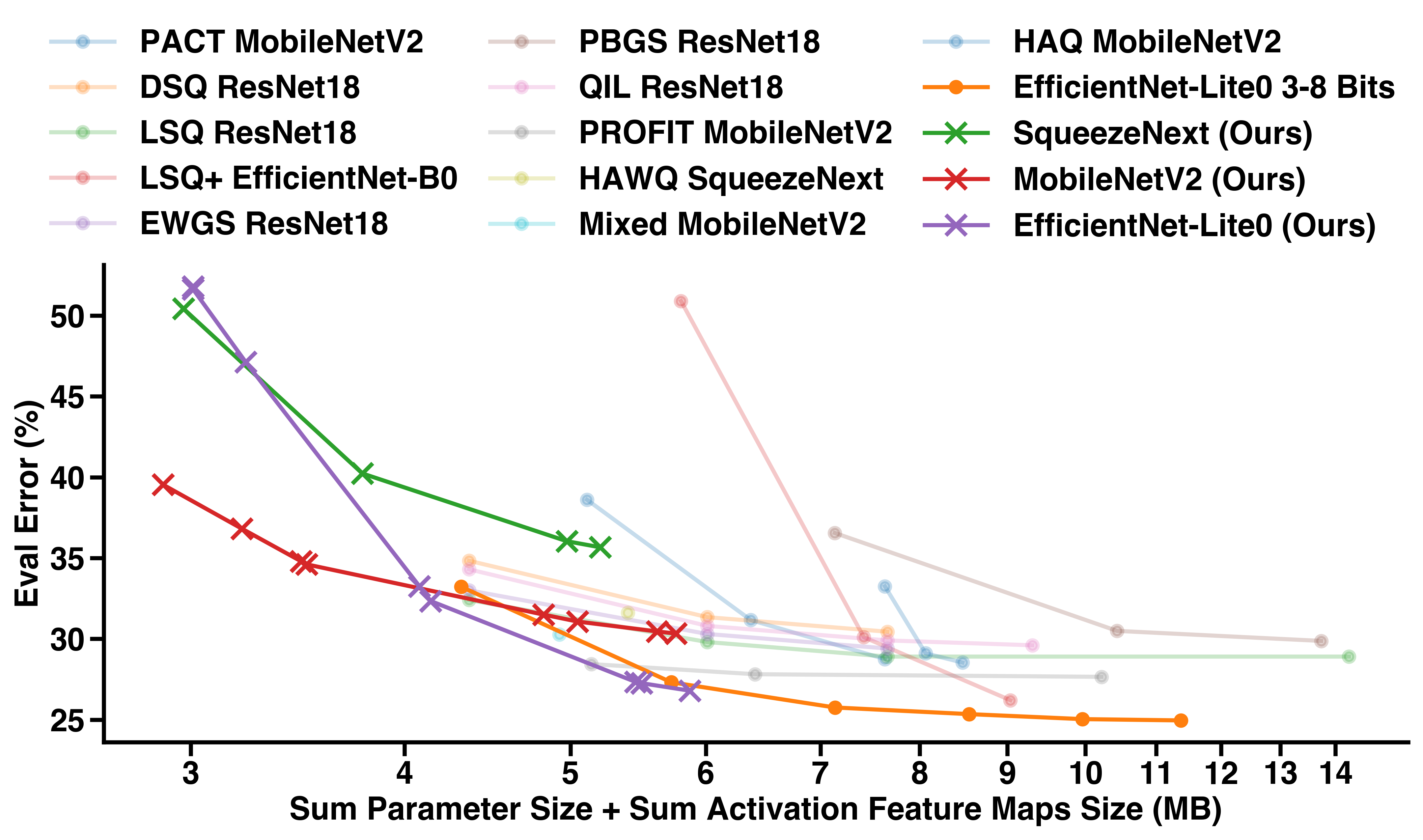}
  \caption{Our results quantizing different models compared to state-of-the-art. Network size (sum of parameters and activations) is compared to the evaluation error on the ImageNet dataset. Models quantized with our method occupy the Pareto-frontier, delivering smaller multi-bit networks at higher accuracy. We shows results from: PACT\cite{choi2018pact}, DSQ\cite{gong2019differentiable}, LSQ\cite{esser2019learned}, LSQ+\cite{bhalgat2020lsq+}, EWGS\cite{lee2021network}, PBGS\cite{kim2020position}, QIL\cite{jung2019learning}, PROFIT\cite{park2020profit}, HAWQ\cite{dong2020hawq}, HAQ\cite{wang2019haq}, Mixed\cite{uhlich2019mixed} }
  \label{fig:overview}
\end{figure*}

Uniform quantization is emulated in software using rounding and clipping of floating-point values, expressed as:

\begin{equation} \label{eq:1}
    Q_u(x, d, q_{+}) = \mathrm{round}\left( \mathrm{clip}\left(\frac{x}{d}, - q_{+}, q_{+} \right)  \right) \cdot d.
\end{equation}

Where $Q_u$ is the quantization function, $x$ the value to be quantized, $d$ the step size and $q_{+}$ is the dynamic range. Achieving high model-accuracy given $b$ bits to represent values, entails a careful choice of the dynamic range ($q_{+}$) and (implicitly) the step size ($d$). The dynamic range relates to the step size and the number of bits as $b = \log (q_{+}/d)+1$. In this paper, we use \textit{calibration} for the process of determining these values during training. As part of calibration, $q_{+}$ is often smaller than $\mathrm{max}(\mathrm{abs}(x))$ to further improve quantization efficiency. See supplementary materials~\ref{calib-text} for a detailed study on the different scaling between $\mathrm{max}(\mathrm{abs}(x)$ and $q_{+}$.

There are two main approaches to quantizing a floating point model: Post training quantization (PTQ) and Quantization Aware Training (QAT). PTQ does not typically require retraining or fine-tuning, with recent work by Dai et al.~\cite{dai2021vs} demonstrating a 6 bit quantized ResNet-50 delivering 75.80\% accuracy on ImageNet. However, the authors report limited success in but retaining accuracy in low-bit regimes, dropping from 75.80\% with 6 bits to 7.11\% with 3 bits. QAT takes into account quantization of weights and activations during training, and consequently has seen greater use in the low-bit regimes. However, QAT must address challenges arising from the non-differentiability of the quantization function which results in vanishing when propagated through multiple quantized layers.

\subsection{Quantization Aware Training}
%Quantizing a floating point model without retraining or fine-tuning is called post training quantization (PTQ). PTQ techniques can either be entirely data-free or use some calibration data. In a study of different calibration methods~\citet{dai2021vs} demonstrated 75.80\% evaluation accuracy on ImageNet using a ResNet50 quantized down to 6 bits. They devise a more fine-grained, per-vector calibration technique where each matrix is broken down into smaller vectors which are independently calibrated. As a calibration technique for $q_{max}$~\citet{dai2021vs} recommend using the $99.99^\mathrm{th}$ quantile computed from a subset of the training data. Although PTQ can quantize models with limited access to training data and no knowledge about the training process, retaining accuracy in the low bit-width regimes remains a challenge, e.g. for 3 bits~\cite{dai2021vs} performance degrades to 7.11\%. Addressing this challenging low bit-width regime researchers exploit the inherent resilience and adaptation of NN models and apply a technique called quantization aware training (QAT). QAT quantizes activations and weights with access to the training process, adapting the model to the effects of quantization. The main hurdle in doing so is that the rounding function for quantization is non-differentiable,

The straight-through-estimator (STE)~\cite{bengio2013estimating} is commonly used to avoid this problem, replacing the derivative of a discretizer (rounding) with that of an identity function. Essentially, ignoring the rounding function in the backward pass and preserving gradient flow.

To further enhance QAT performance, Choi et  al.~\cite{choi2018pact} introduce PACT which makes the dynamic range (see eq.~\eqref{eq:1} $q_{+}$) a trainable parameter. They achieve 75.3\% evaluation accuracy on ImageNet with a ResNet50 quantized down to 3 bits while simultaneously stabilizing training. Jung et al.~\cite{jung2019learning} examine the use of non-uniform quantization and study the effect of learned quantization levels. The resulting quantizer achieves 73.1\% evaluation accuracy on ImageNet while using a ResNet34 quantized to 3 bits. However, non-uniform quantizers cannot always be mapped on to fixed-point arithmetic and can incur significant overhead when deployed or implemented in hardware~\cite{dong2020hawq}. Learning the step-size $d$ while learning a uniform quantization scheme, Esser et al.~\cite{esser2019learned} developed LSQ which achieves an accuracy of 74.3\% on ImageNet using a 3-bit ResNet34. In LSQ+, Bhalgat et al.~\cite{bhalgat2020lsq+} build on this technique by parameterizing the symmetry of the quantizer to accommodate modern activation functions such as swish, h-swish and mish, which have limited, but critical negative excursions. They demonstrate the effectiveness of their method on modern architectures by achieving 69.9\% accuracy with a 3-bit Efficient-B0 and 66.7\% accuraacy with a 3-bit MixNet on ImageNet. However, these methods do not quantize the first and last layers of the models, which typically incur significant performance degradation. In contrast, recent work using progressive-freezing and iterative training to quantize MobileNets~\cite{park2020profit} achieve 71.56\% accuracy on ImageNet (note that MobileNets are significantly smaller than ResNets).

\subsection{Heterogeneous Quantization}

Model accuracy is not equally sensitive to quantization in different layers~\cite{park2020profit}.  Prior work leverages this to allocate numerical precision on a per-layer basis ~\cite{yao2020pyhessian, park2020profit, wang2019haq, elthakeb2018releq}. In HAQ~\cite{wang2019haq}, the authors use reinforcement learning with a hardware simulator generating energy and latency estimates to optimize the bit-width of every layer. They report a $1.9\times$ improvement to energy and latency while maintaining 8-bit levels of accuracy for MobileNet models. More recently, second order techniques like~\cite{dong2020hawq} use Hessian eigenvalues as a quantization sensitivity metric and assign layer bit-widths. This Hessian aware trace-weighted quantization (HAWQ) offers 75.76\% accuracy (ImageNet) for ResNet50 with an average of 2 bit weights and 4 bit activations.

Uhlich et al.~\cite{uhlich2019mixed} (label Mixed in Figure~\ref{fig:overview}) formulate a fully differentiable quantization scheme, where both the step-size and the dynamic range are trainable, using a symmetric uniform quantizer $Q_U(x, d, q_{+})$. This formulation implicitly learns the bit-width. Additionally, the authors add an additional constraint to the loss function to target a network weight size and maximum feature map size. %The constraint and quantizer both use straight-through estimators for rounding functions in the backward pass.
Taken together, their improvements result in a MobileNetV2 with a weight memory footprint of 1.55~MB and a maximum activation feature size of 0.57MB, while delivering an accuracy of 69.74\% on ImageNet (an estimated 4.93 MB for the weights and sum of activations).

\subsection{Gradient Scaling for Quantization}
% Recent work indicates that the  discrepancy in gradients between the rounding function and STE  limits model performance in the ultra-low bit-width regimes (1-3 bits)~\cite{lee2021network}. They propose modifying the gradient updates to  account for this discrepancy, reporting \red{XYZ} performance. 
% I do not know which other work you meant here... sorry

Most models employ straight-through-estimators (STEs) in QAT for quantized neural networks essentially ignoring discretization in the backward pass. Alternatively some recently proposed techniques avoid this discrepancy by employing a smooth function, e.g., stacked $\tanh$, prior to the non-differentiable quantizer to emulate quantization effects facilitating gradient flow across layers (DSQ~\cite{gong2019differentiable} in Figure~\ref{fig:overview}). They demonstrate a 2-bit ResNet18 delivering 65.17\% accuracy on ImageNet. Since the authors cascade this soft quantizer with a hard quantizer, they still employ an STE to propagate gradients in the backwards pass. %The authors still use a STE for the actual quantization in the backward pass, only inserting an additional smooth function prior to quantization.

Kim et al.~\cite{kim2020position} (PBGS in Fig.~\ref{fig:overview}) propose gradient scaling as a regularizer to learn `easy to quantize' networks, training models with gradients scaled to induce values on the quantization grids. The authors demonstrate results for a 4-b quantized ResNet18 trained to  63.45\% evaluation accuracy on ImageNet. Nguyen et al.~\cite{nguyen2020quantization} achieve the same effect through regularization, using the absolute cosine function for a 6-bit automatic speech recognition recurrent neural network with only a 2.68\% accuracy degradation compared to the baseline model. Lee et al.~\cite{lee2021network} (EWGS in Figure~\ref{fig:overview}) combine quantization in the forward pass and gradient scaling in the backward pass to account for discretization errors between inputs and outputs of the quantizer. They incorporate the sign and magnitude of the discretization error as well as second-order information to determine the gradient scaling factor. This method achieves 67\% evaluation accuracy on ImageNet with ResNet18 quantized to 2 bits.% (a model with a memory footprint of 4.36 MB including wgts. and sum of acts.).

\section{Methods}\label{sec:meth} 
%a training phase to find the optimal bit-widths employing penalty scheduling and alternating quantizer parameter updates, and 

Given a pretrained floating point model we quantize it in three phases: (i) homogeneous pre-training phase, (ii) a phase to learn precisions, and (iii) a final finetuning phase where only model parameters are updated (see Fig.~\ref{fig:quantphase}). During all those phases we employ gradient scaling to account for discretization in the backward pass.  We implement an initial calibration phase for both weights and activations using Gaussian calibration for the weights and the $99.99^\mathrm{th}$ percentile for the activations, which improves our homogeneous quantization performance by up to 1.22\% for 3-bit EfficientNet compared to sample maximum calibration (see suppl. mat. Table~\ref{appendix:calib} for details). During phase (ii) we employ penalty scheduling and reduce the frequency of quantizer parameter updates to combat learning instabilities. We also incorporate the weight and activation size penalty into the loss function like Uhlich et al.~\cite{uhlich2019mixed}, resulting in:

\begin{equation} \label{eq:2}
\begin{aligned}
   L = & CE(x, y) + \beta \max \left( \left( \sum^L_{l=1} \sum^C_{c=1} b^w_{lc} \cdot s_{lc}^w \right)  - t^w ,0 \right)^2\\
   & + \beta \max \left( \left( \sum^L_{l=1} b^a_l \cdot s_l^a\right) - t^a, 0\right)^2.
\end{aligned}
\end{equation}

% new figure/table with three panels, what is the effect for the elements in your recipe. e.g. when i not do (just a couple of runs) -> this particular set of steps is what matters. Findings in introduction novelty in that paper. 

% in this paper across multiple runs its not that different, new findings point them out in introdution.

% why should people read your paper.

Here, $x$ is the input, $y$ the target, $CE$ stands for the cross entropy loss, $b_l$ represents the bitwidth of layer $l$, $s_l$ is the number of parameters of a given layer (the additional $c$ indicates the channel) and $t$ is the target size for the model. The superscripts $w$ and $a$ indicate weights and activations respectively. We use rectified quadratic penalties to enable an accuracy vs. model size trade-off during training, with the rectification used to prevent penalization once the size budget is met. A single modulating factor $\beta$ controls the penalty on model size.

We use the sum of weights and activation feature maps as our cost metric for efficiency. Data movement caused by memory accesses for weights and activations dominates other factors in modern edge accelerators~\cite{wu2019accelergy}. Consequently, for such hardware, accounting for the total number of memory accesses for heterogenously quantized tensors, is crucial to determining model efficiency. Existing metrics like number of operations or parameter footprint, do not fully consider data movement or variable precision tensors. The Arithmetic Computation Effort (ACE)~\cite{zhang2022pokebnn} is an alternative compute-focused metric that has been proposed for multi-precision edge accelerators. Our metric is strongly correlated with ACE (0.956) across multiple configurations, but offers more direct interpretability of the cost. %use a metric called Arithmetic Computation Effort (ACE) which has a 0.9563 correlation coefficient to our metric and we believe that our metric is more accessible to a broader audience. % We are not using number of parameters or number of operations because they would not capture the effect of reduced/mixed precision.

\begin{figure}
  \centering
  \includegraphics[width=0.75\columnwidth]{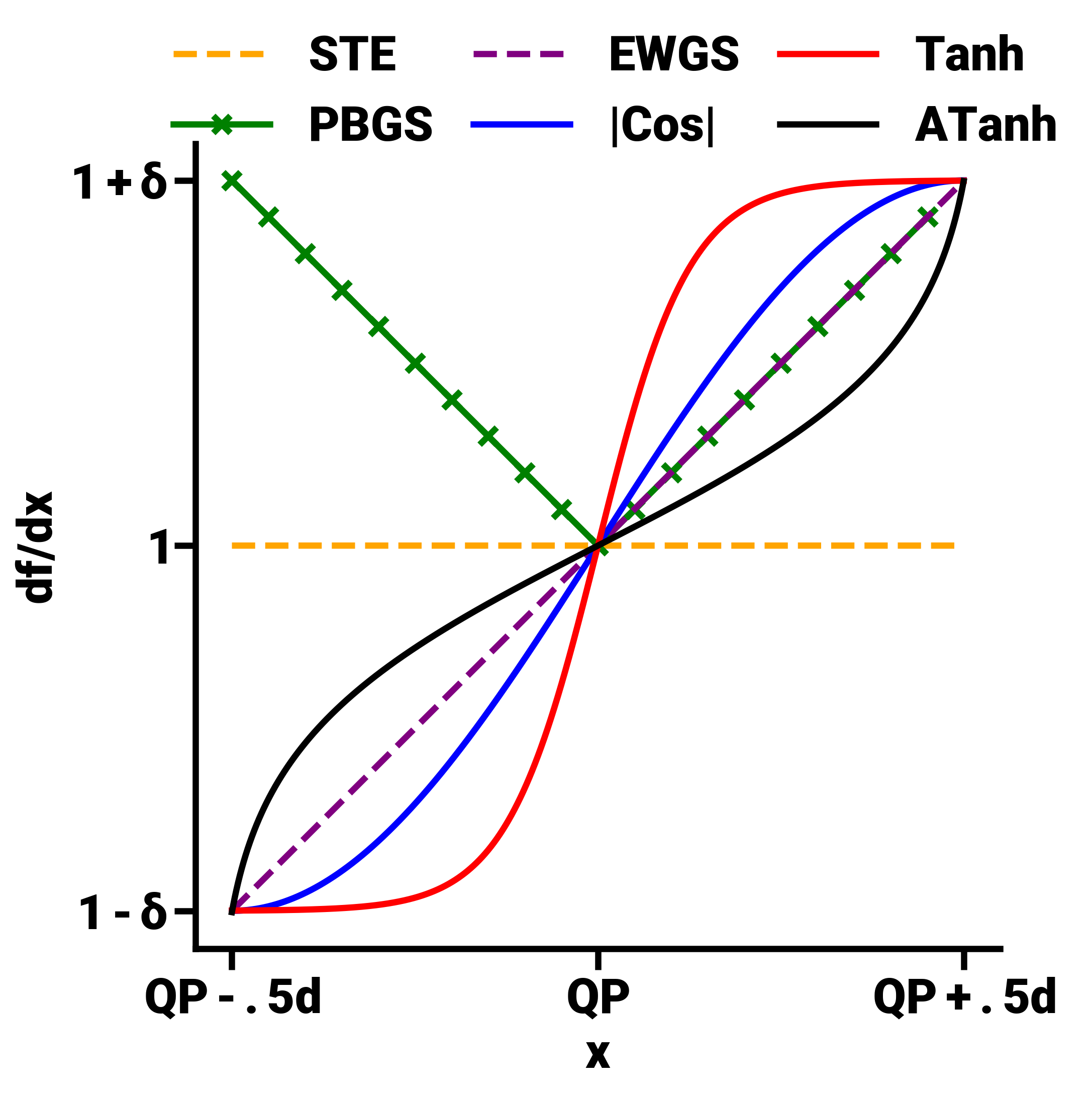}
  \caption{We illustrate different gradient scaling functions: straight-through-estimators (STE~\cite{bengio2013estimating}), elementwise gradient scaling (EWGS~\cite{lee2021network}), position based gradient scaling (PBGS~\cite{kim2020position}), absolute cosine regularization (Acos~\cite{nguyen2020quantization}) as well as hyperbolic tangent function (Tanh) and its inverse (InvTanh). Note that $QP$ denotes quantization point, $d$ is the step size and $\delta$ is the magnitude control hyper parameter for gradient scaling.}
  \label{fig:2}
\end{figure}

% \begin{figure*}
%   \centering
%   \includegraphics[width=0.75\textwidth]{figures/meth_sur.png}
%   \caption{On the left we illustrate different gradient scaling functions: straight-through-estimators (STE~\cite{bengio2013estimating}), elementwise gradient scaling (EWGS~\cite{lee2021network}), position based gradient scaling (PBGS~\cite{kim2020position}), absolute cosine regularization (Acos~\cite{nguyen2020quantization}) as well as hyperbolic tangent function (Tanh) and its inverse (InvTanh). Note that $QP$ denotes quantization point, $d$ is the step size and $\delta$ is the magnitude control hyper parameter for gradient scaling. On the right we show the performance of mixed gradient scaling functions (e.g. different gradient scaling functions for weights and activations) on a homogeneous 3 bit EfficientNet-Lite0 on the ImageNet dataset. }
%   \label{fig:2}
% \end{figure*}

% \subsection{Pretraining}

\subsection{Quantization Training Dynamics}\label{sec:handcrafted}

\begin{figure*}
  \centering
  \includegraphics[width=0.8\textwidth]{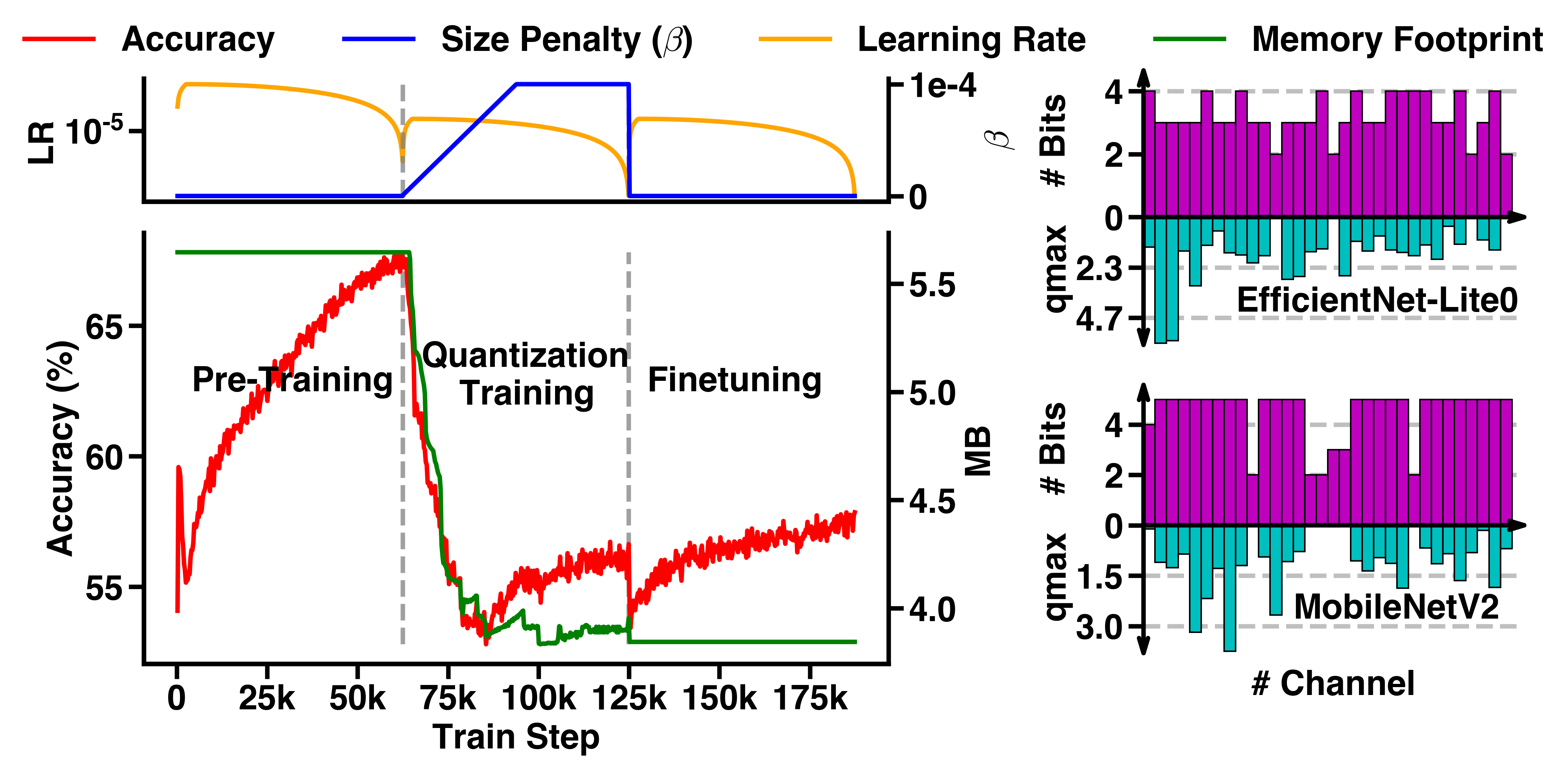}
  \caption{On the left illustrating the three phases of our mixed precision training method. The top left shows scheduled learning rate and size penalty ($\beta$) term meanwhile the bottom left shows the evolution of model accuracy and model size. On the right we show an example per-channel bit allocation in a weight kernel of an EfficientNet-Lite0 and MobileNetV2. Extreme quantization is correlated to low dynamic range ($q_{+}$). However, the contra does not necessarily hold.}
  \label{fig:quantphase}
\end{figure*}

Learning both model parameters and bit-width allocations is necessarily a higher dimensional problem than just learning the parameters alone. This increases the search's sensitivity to initial conditions. To mitigate this, we de-couple these two elements by starting out training on a homogeneously quantized network (from the pre-training) and then training the per-layer quantizer with progressively increasing pressure from the model size constraints. As shown in Figure~\ref{fig:quantphase} (left), the model suffers from dramatic accuracy degradation when the model size penalty is imposed. Often, this results in catastrophic training failure due to the model size penalty dominating cross-entropy loss. We use a soft-transition on the size penalty ($\beta$) by linearly increasing $\beta$ to it's final value. This initial training can be viewed as enabling coarse navigation to the optimal in the solution space, followed by a more fine-grained descent towards the joint model-quantizer optima. As seen in Figure~\ref{fig:quantphase} (left), parameter updates can recover some lost accuracy after $\beta$ saturates. 

% We set the initial relaxed budget which is then progressively tightened

We observed model instability and training failure when both bit-precision and model parameters were updated frequently. Infrequent updates prevented thorough exploration of the model search space, resulting in models that were still similar to their homogeneously quantized initializations. We avoid this by limiting the bit-precision update frequency, restricting updates to every $\Phi$ steps. In our experiments, an update frequency of $\Phi=20$ provided sufficient time for model statistics (e.g. batch norm) to stabilize and model parameters to adapt to the new precision level.

Subsequently, we enable finer-grained quantization by adapting the precision of the convolution weight kernels at a per-output-channel granularity (tensor slices). Most existing hardware with mixed-precision support can compute a single channel at a given precision. However, finer grained quantization can entail significant hardware overhead~\cite{ibrahim2021survey}. Banner et al. \cite{banner2019post} have previously shown flexible per-channel bit-widths by solving a layer-wise noise minimization problem, in contrast we learn the per-layer bit-widths based on the overall model loss function. Indeed, the variation in dynamic range across channels from a heterogeneously quantized EfficientNet-Lite0 and MobileNetV2 demonstrate the efficiency gains available using this technique (See Figure~\ref{fig:quantphase} right). To prevent the gains from quantized operation getting negated, we ensure that quantization granularity is not too fine-grained. After these quantizer parameters converge, we freeze them by setting $\beta=0$ in eq.~\eqref{eq:2}. This is followed by an additional fine-tuning period with a decaying learning rate to recover accuracy. In particular, our results suggest that batch normalization statistics are stabilized through this fine-tuning period, recovering accuracy.

\subsection{Gradient Scaling}
The STE operator is the de facto standard for enabling backpropagation through non-differentiable functions. We study how different gradient scaling techniques compare to STE across models and levels of precision. As shown in Figure~\ref{fig:98}, $\tanh$-based gradient scaling for activations and linear scaling for weights~\cite{lee2021network} outperforms other combinations. We examine the effect of gradient scaling across multiple scaling functions ($f\left(x\right)$) enumerated below (and shown in Figure~\ref{fig:2} (left)):
% Our results summarizing the
\begin{enumerate}%[label=(\alph*)]

\item Position based gradient scaling (PBGS)~\cite{kim2020position}:\vspace{-5pt}
\begin{equation*}
\mathrm{scale} = 1 + \delta \cdot |x - \mathrm{round}(x)|.
\end{equation*}

\item Element-wise gradient scaling (EWGS)~\cite{lee2021network}:\vspace{-5pt}
\begin{equation}
\mathrm{scale} = 1 + \delta \cdot \mathrm{sign}(g_x) \cdot (x - \mathrm{round}(x)). \nonumber
\end{equation}

\item Modified absolute cosine (Acos)~\cite{nguyen2020quantization} gradient scaling: \vspace{-5pt}%element-wise gradient scaling (EWGS)
\begin{equation}
\mathrm{scale} = 1+ \delta \cdot \sin(\pi \cdot (x - \mathrm{round}(x)) ). \nonumber
\end{equation}

\item Hyperbolic tangent (Tanh) gradient scaling:\vspace{-5pt}
\begin{equation}
\mathrm{scale} = 1 + \delta \cdot \mathrm{sign}(g_x) \cdot \mathrm{tanh}( \alpha  \cdot (x - \mathrm{round}(x))). \nonumber
\end{equation}

\item Inverse hyperbolic tangent (InvTanh) gradient scaling:\vspace{-5pt}
\begin{equation}
\mathrm{scale} = 1 + \delta \cdot \mathrm{sign}(g_x)   \cdot \mathrm{arctanh}( \alpha \cdot (x - \mathrm{round}(x))). \nonumber
\end{equation}

\end{enumerate}

Here, $\delta$ is a general hyperparameter to modulate the magnitude of gradient scaling and $g_x$ is the gradient. We also introduce an additional hyperparameter, $\alpha$, to control the steepness of the hyperbolic tangent functions. Figure~\ref{fig:98} (right) shows the performance of various gradient scaling techniques for activations and weights when homogeneously quantizing EfficientNet-Lite0 to 3 bits. We examined the effect over 20 trials and show the resulting distribution in Figure~\ref{fig:98}. The horizontal dotted lines represent the baseline performance (when both acts. and wgts. use the same scaling technique) of the STE and EWGS method respectively.  We observe that different gradient scaling schemes benefit the training performance for weights and activations, owing in part to different underlying distributions.  Indeed, employing STEs for both weights and activations provides the performance baseline. We note that the improvements derived from gradient scaling on activations only (configurations \textit{a}, \textit{b}, \textit{c}) lead to lower performance gains when compared to applying gradient scaling on weights only (configurations \textit{d}, \textit{e}, \textit{f}). This suggests that the performance gains from gradient scaling can be primarily attributed to gradient scaling of weights. Although configuration \textit{e} (weights use EWGS and activations use STE) delivered the best single-run result, as seen in Figure~\ref{fig:98}, these results were not consistent across trials. Indeed, in some trials, this configuration performed worse than the baseline. Indeed, the majority of the performance gains from applying EWGS area also observed in configuration \textit{e}, where EWGS is only applied to weights while the activations use STEs. We observed that linear gradient scaling (EWGS) for weights and the inverse of the hyperbolic tangent scaling of activation gradients provides consistent improvement over the baseline, proving to be the more robust gradient-scaling technique. More comprehensive sensitivity analysis and analysis of computational overhead are provided in~\ref{asec:grad_scale} in the supplementary material.

% the propagation of quantization errors through multiple activation layers from benefits from .%For the performance of uniform gradient scaling (e.g. same for weights and activations) see Figure~\ref{fig:uniform} in the supplementary materials. \sj{we also show 2 and 4b results in suppl. mat. but you're not referencing that?}

% compare against ACE other cost metrics section

\section{Experiments}

We demonstrate the effectiveness of our proposed recipe on the ImageNet~\cite{deng2009imagenet} dataset across multiple models including EfficientNet-Lite0~\cite{tan2019efficientnet}, MobileNetv2~\cite{sandler2018mobilenetv2}, wide SqueezeNext~\cite{gholami2018squeezenext}, and ResNet18~\cite{he2016deep}. We use the Flax~\cite{flax2020github} and Jax~\cite{jax2018github} frameworks to implement the networks, quantization scheme, and training routine. All codes are available under \url{https://github.com/Intelligent-Microsystems-Lab/HeterogeneousQuantization}

\subsection{Setting}
We ran our experiments on TPUv3 accelerators using the Google Cloud. Each experiment ran on a single instance which comprises 8 cores and 32GB memory. We provide some hardware synthesis-based estimates of the latency impact of quantized models in Supplementary Materials~\ref{app-lat}.

\begin{figure}
  \centering
  \includegraphics[width=0.8\columnwidth]{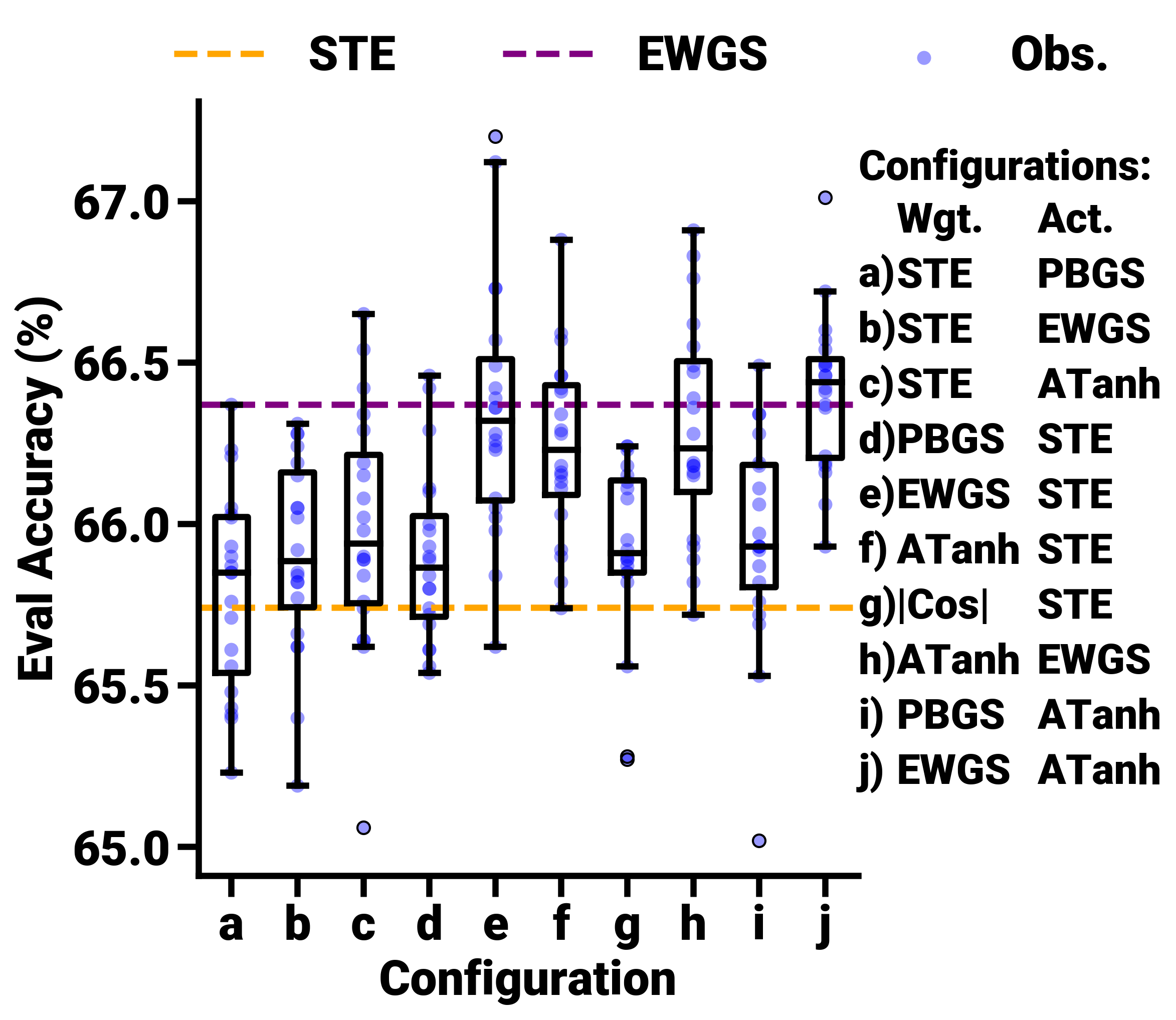}
  \caption{The performance of different mixed gradient scaling functions (different gradient scaling for weights and activations) on a homogeneous 3 bit EfficientNet-Lite0 on ImageNet.}
  \label{fig:98}
\end{figure}

We implement standard input pre-processing for training, randomly cropping images to $224 \times 224$ with 3 channels (RGB), followed by input augmentation. Our augmentations include a random flip (left or right) and channel normalization (mean 127 and standard deviation 128). During evaluation the image is cropped around the center and no random flip is applied. Training uses RMSProp~\cite{optax2020github} with $0.9$ Nesterov momentum and a learning rate of $10^{-4}$. We increase the learning rate linearly from zero during the first two epochs and subsequently reduce it to zero in a cosine decay. Our training batch size is 1024, which is evenly split across the 8 cores of the TPU for single-program, multiple-data (SPMD) parallelism. We apply weight decay of $10^{-5}$ and label smoothing of $10^{-1}$ for improved accuracy. Each QAT training phase (pretraining, heterogenous training, and finetuning) lasts 50 epochs.

\subsection{Gradient Scaling}

We determine the gradient scale factor $\delta$ ($5\cdot10^{-3}$) through a grid search. Results of the gradient scaling can be seen in Figure~\ref{fig:98} (right) and for non-mixed configurations in supplementary materials~\ref{asec:grad_scale}. Both show data from 20 trials on a 3-bit homogeneously quantized EfficientNet-Lite0, illustrating the variance in accuracy from employing gradient scaling methods. Figure~\ref{fig:gradscale-sweeps} (supplementary materials) shows a smaller grid search for ideal gradient scaling on the same network quantized to 2 and 4 bits. Notably, variance in final accuracy increases when quantizing to fewer bits whereas the difference in accuracy gets exacerbated.

\begin{figure*}
  \centering
  \includegraphics[width=.7\textwidth]{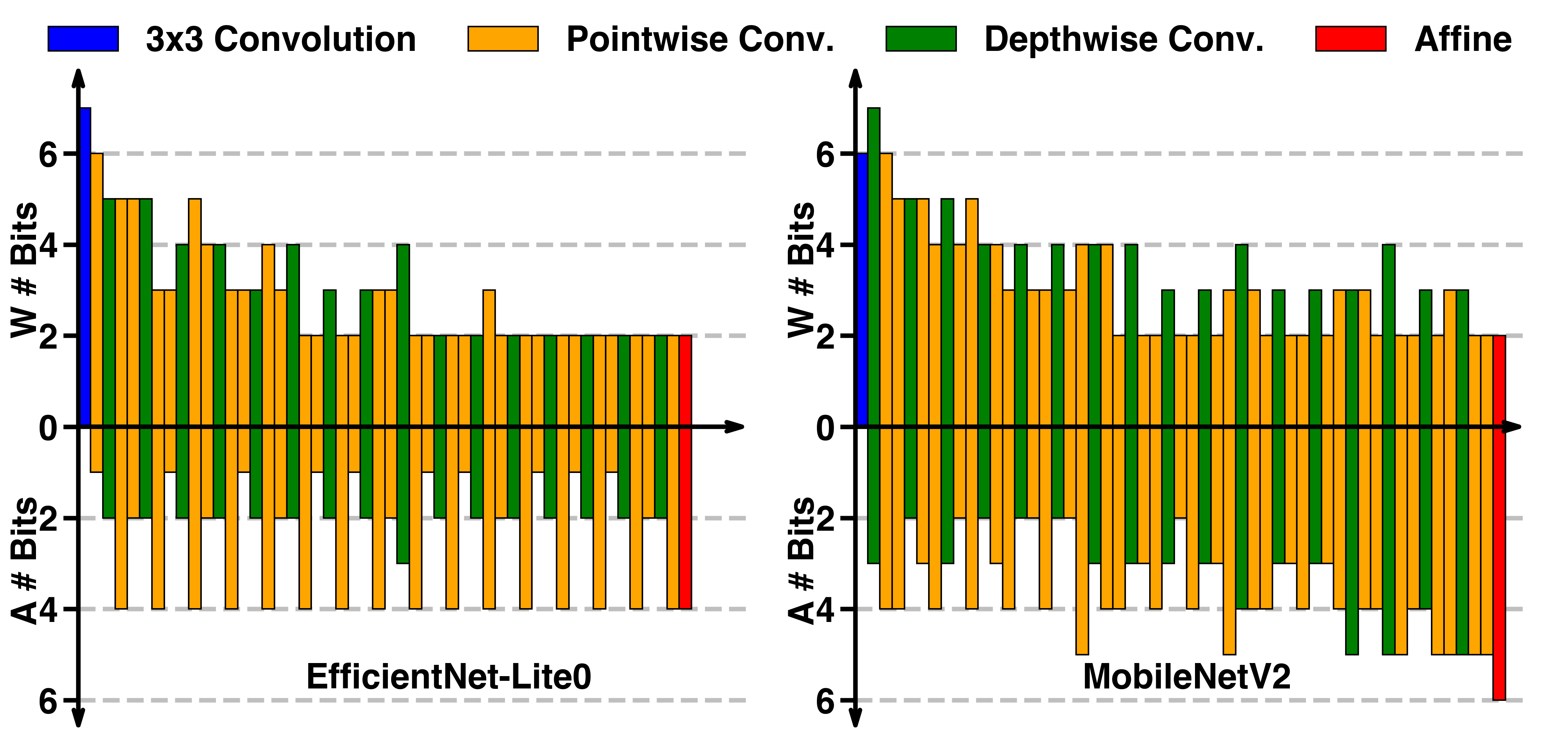}
  \caption{Internal bit allocation across layers of weights (up) and activations (down) for EfficientNet-Lite0 and MobileNetV2. Weights in the first layers have higher bit-widths for both models. Activations bitwidths for EfficientNet-Lite0 form a high precision path, e.g. activations which are residuals have higher precision. For both models the last affine layer has high precision.}
  \label{fig:inside}
\end{figure*}

\subsection{Model Quantization}

Figure~\ref{fig:overview} provides context for our results, comparing our techniques with state of the art quantization methods~\cite{choi2018pact, gong2019differentiable,esser2019learned,bhalgat2020lsq+,lee2021network,kim2020position,jung2019learning,park2020profit,dong2020hawq,wang2019haq,uhlich2019mixed}. If unavailable, we computed the weight and activation sizes based on our replication of their work (all assumptions are shown in the supplementary materials Table~\ref{tab:asum:}). The total network size (x-axis in Fig.~\ref{fig:overview} includes batch norm parameters for which we assume a bfloat16 data type. Although no small-scale model (including our own) reports quantized batch normalization, some related work do report results from quantizing the batch norm layers for a 4-bit ResNet18 with a higher memory footprint than our reported results Yao et al.~\cite{yao2021hawq}.

Our method improves the Pareto frontier for model error and model size, with the greatest improvement seen in the sub 6 MB region. Here, we define size to be the sum of weights and sum of activation feature maps. We specifically optimize the total size of the activations, because of the high energy cost arising from data movement for activations~\cite{chen2019eyeriss, yazdanbakhsh2021evaluation}. To the best of our knowledge, we report  best-in-class accuracy for multi-bit models in extreme constraint settings with less than 4.3 MB available for both weights and activations. Our efficient frontier (Figure~\ref{fig:overview}) shows the degradation in performance across various target budgets and encapsulates the difference in robustness of various quantized model architectures. The EfficientNet-Lite0 architecture performs well for homogeneous quantization and heterogeneous quantizaton in a range between 4-6 MB total model size, meanwhile MobileNetV2 delivers better accuracy below 4 MB. We note the different scaling trends seen for quantized EfficientNet-Lite0 and MobileNetV2, suggesting that MobileNetV2 may be more suitable for ultra-low budget applications. We included a wide SqueezeNext model due to its small weight footprint, however due to its depth the activation size dominate the overall model budget and making it a strictly worse model across the target budgets. ResNet18 and SqueezeNext results are provided in~\ref{asec:morers}. % in the supplementary materials.

% For our experiments we pick the homogeneous quantization calibration method through a grid search trying out different combinations (of weight and activation calibration) of popular calibration choices. The results can be seen in supplementary materials Table~\ref{appendix:calib}. 
% yes this is still a repeat of stuff from the method section, should I try to move it up?

% Indicating the bit-widths for weights on the top and for activation on the bottom while different types of layers () have different colors.
Figure~\ref{fig:inside} shows the detailed layer-wise bit-allocation of heterogeneously quantized EfficientNet-Lite0 with a total size of 3.43 MB and MobileNetV2 with 3.41 MB. The EfficientNet-Lite0 architecture imposes greater precision in weights at the early layers of the network compared to the later layers. However the activations follow a constant pattern throughout the network. Notably, the activation of the last pointwise convolutional layer of a mobile inverted bottleneck (MBConv block) are higher in comparison to the other layers of MBConv blocks. Our results suggest that pointwise convolutional layers that have both residual and direct inputs require much higher precision to prevent quantization-induced information loss. This high precision bit-allocation indicates a critical information flow pathway. The MobileNetV2 bit allocation is similar to that of EfficientNet-Lite0, with higher precision weights in the initial layers which reduce with network depth. The critical path for the activation is not as pronounced as it is for the EfficientNet architecture and additionally high activation bit-width are allocated to layers closer to the final affine layer whose activation bit-width is the highest.

\subsection{Additional Consideration}

\subsubsection{Bias Quantization}
While our reported models quantize biases, we also examined the effect of keeping biases at higher bfloat16 precision. Typically, accelerators can pre-load biases into the hardware accumulator, minimizing the energy impact. We summarize the impact of bias quantization in Table~\ref{tab:distill}. Crucially, we note that MobileNetV2 at larger memory budgets do not see accuracy benefits ($\le 0.01\%$ higher accuracy). However, EfficientNet-Lite0 models with tight memory budgets do see an increase their accuracy by approx. 1.56\%.

\subsubsection{Knowledge Distillation}

\begin{table*}
  \caption{Effect of knowledge distillation (KD) on heterogeneously quantized networks and unquantized biases for the last affine layer. The first column shows the effect of KD on floating point networks the following columns are networks from the efficient frontier. Soft-label KD refers to a KD technique where the targets of the student network are the predictions of the teacher network~\cite{hinton2015distilling}. Meanwhile penalty KD uses one-hot encoding as the target and adds a penalty term to force the student model prediction to align to the teacher~\cite{park2020profit}.}
  \label{tab:distill}
  \centering
\begin{tabular}{lrrrrrrrr}
\toprule
\multicolumn{9}{c}{EfficientNet-Lite0}                                                                 \\
\midrule
Size (MB)     & 22.66  & 3.01  & 3.23  & 3.98  & 4.14  & 5.45  & 5.50  & 5.87  \\
Base Accuracy (\%)      & 75.53 & 48.37 & 52.87 & 66.46 & 67.66 & 72.56 & 72.75 & 73.21 \\
Soft-Label KD & -0.31  & 0.21  & 0.24  & 0.11  & 0.10  & 0.17  & 0.10  & -0.11 \\
Penalty KD    & -0.22  & 0.16  & 0.27  & 0.08  & 0.16  & 0.17  & 0.04  & -0.11  \\
\midrule
No Bias Quant & -  & 1.56 &	1.16 &	0.20 &	0.26 &	0.14 &	0.15 &	0.00 \\
\bottomrule
\vspace{.45em}
\end{tabular}

\begin{tabular}{lrrrrrrrrr}
\toprule
\multicolumn{10}{c}{MobileNetV2}                                                                        \\
\midrule
Size (MB)     & 20.25 & 2.89  & 3.21  & 3.48  & 3.51  & 4.82  & 5.05  & 5.62  & 5.76  \\
Base Accuracy (\%)      & 71.46 & 60.72 & 63.18 & 65.20 & 65.39 & 68.50 & 68.93 & 69.54 & 69.68 \\
Soft-Label KD & 0.15  & 0.36  & 0.12  & 0.13  & 0.23  & 0.12  & -0.26 & -0.24 & 0.02  \\
Penalty KD    & 0.03  & 0.16  & 0.17  & 0.17  & 0.28  & 0.11  & -0.19 & -0.23 & -0.03 \\
\midrule
No Bias Quant & - & 0.38 &	-0.01 &	0.24 &	0.08 &	0.21 &	0.05 &	0.16 &	0.13 \\
\bottomrule
\end{tabular}
\end{table*}

Recently proposed quantization techniques have shown that applying knowledge distillation (KD) to their existing quantization techniques can improve results. We examine how KD impacts the EfficientNet-Lite0 and MobileNetV2 models on our Pareto frontier~\ref{fig:overview}. We use the KD process in~\cite{hinton2015distilling}, using soft labels created by a B16 vision transformer~\cite{chen2021vision} with an accuracy of 85.49\% (soft-label KD). We also examined the knowledge distillation technique employed by PROFIT~\cite{park2020profit}, here the knowledge of the teacher model is induced into the student model through a penalty term in the loss function (penalty KD). The results summarized in Table~\ref{tab:distill} show that knowledge distillation can have a positive effect on the accuracy of up to 0.36\% but can also have negative effects. Neither of the KD techniques examined dramatically altered the model accuracy across the Pareto frontier.

\subsubsection{Exploring Training Schedule and Quantization Approaches}

We evaluate our proposed QAT schedule against both automatically searched schedules and those derived using convex optimization approaches. %derived using automated techniques and validate its wide applicability. We also illustrate the efficacy of our gradient-based learning of the quantization parameters compared to alternative

% To validate our method, we directly compare our schedule and quantization approach to alternatives as detailed below.

\paragraph{Automated Schedule Search}

\begin{figure}
  \centering
  \includegraphics[width=.8\columnwidth]{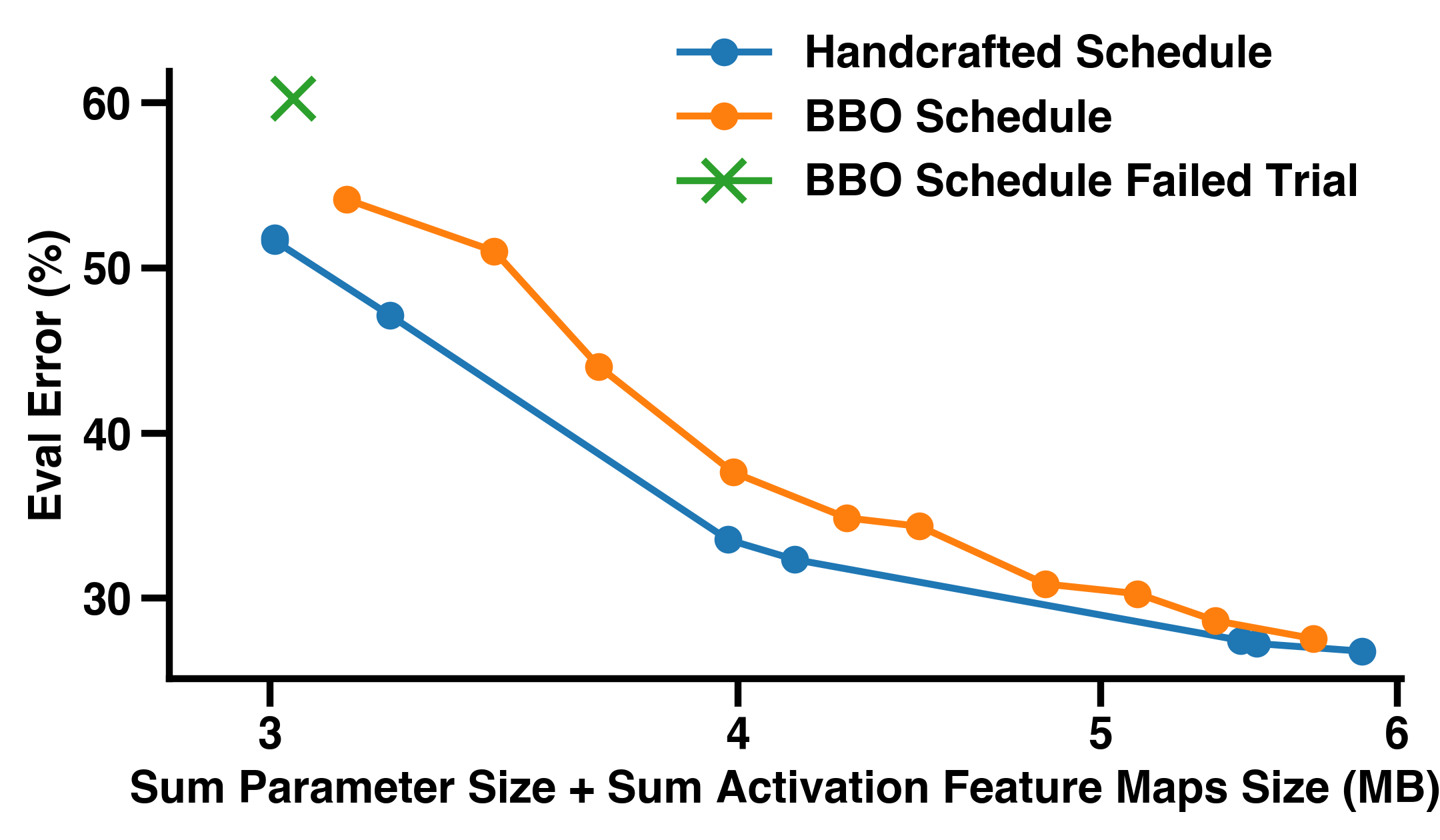}
  \caption{Comparison between QAT employing our schedule and the scheduled derived through a distributed black-box optimization (BBO) method evaluated for quantizing an Efficient-Lite0. The BBO-derived schedule performs strictly worse than ours and fails to train a network on the strictest budget (2 bits on average).}
  \label{fig:bb_sched}
\end{figure}
% black-box T o improve the results potentially; we also used a black-box optimization algorithm to determine a training schedule. We used standard black-box optimization   algorithms
We compare our handcrafted QAT schedule (see Fig.~\ref{fig:quantphase}) to a schedule derived through distributed black-box optimization (BBO)~\cite{oss_vizier,google_vizier}, to determine the performance of our approach. We set up the BBO search space to include: (i) frequency of bit-width update, (ii) weight \& activation penalty ($\beta$ in eq~\eqref{eq:1}), (iii) homogeneous bit-width for pre-training, (iv) ramp-up length of the quantization training, and (v) ramp-up mode (linear, cosine, or exponential). The BBO was directed to maximize accuracy for the EfficientNet model on a randomly sampled subset of the training data while minimizing the discrepancy between model budget and achieved size. We chose an average of 3 bits for the model budget (1731 kB for weights and 2505 kB for the sum of activations). The BBO conducted 766 evaluations within our compute budget (approx. 14,000 accelerator hours) to converge to a recipe. 

Figure~\ref{fig:bb_sched} compares our schedule against the schedule determined by the BBO, showing that our schedule consistently outperforms the automated search. For the strict budget (average of 2 bits for each tensor), the BBO-derived schedule exceeds the allocated budget by 135.15~kB. We hypothesize that our hand-crafted schedule outperforms the BBO schedule due to the large search space, with varying levels of sensitivity to the scheduling parameters.

\paragraph{Optimization-Based Quantization}

Our problem requires simultaneously optimizing for accuracy and model size, constrained by a memory budget (see eq.~\eqref{eq:2}). Alternating direction method of multipliers (ADMM)~\cite{boyd2011distributed}, combines the decomposability of dual ascent with the convergence guarantees of the method of multipliers making it an attractive solution with theoretical grounding. We reformulate our quantization to be compatible with ADMM by separating the problem objectives into optimizing for accuracy and model size, constrained by equality between model parameters between the two optimization steps. ADMM, then operates on two sets of model weights, updating weights optimizing each objective and until the two sets of parameters converge. % The details for this are described in ~\ref{asec:admm}.

Our experimental findings demonstrate the superiority of our gradient-based approach over ADMM. When quantizing an EfficientNet-Lite0 to an average bit-width of 4 bits, our method achieved 72.46\% accuracy, surpassing the 69.21\% achieved by ADMM. Details on the ADMM formulation and parameters (including hyperparameter search) are in ~\ref{asec:admm}.

\section{Conclusions}

We introduce a recipe for quantization-aware training for heterogeneously quantized neural networks where bit-widths are trained alongside model parameters. We employ a novel gradient scaling function to account for discretization due to quantization in the backward pass. Combined with careful scheduling in penalizing the accuracy loss and model size allows us to achieve exceed state-of-the-art model quantization. Models quantized by our technique occupy the Pareto optimal frontier of model size (including weights and activations) against performance (evaluation error) on the ImageNet dataset. To the best of our knowldege, our methods delivers best in class multi-bit neural networks with total memory footprint below 4.3~MB. Extensive evaluation and sensitivty analysis verifies our quantization performance. %Additionally we analyzed the bit allocation of the EfficientNet-Lite0 and MobileNetV2 architecture pointing out critical paths for activations and high bit-width requirements for weights in early layers of DNNs.

% \subsection*{Acknowledgments}
% % \textcolor{purple}{removed for review}
% Research supported with Cloud TPUs from Google's TPU Research Cloud (TRC)

% \bibliographystyle{plainnat}
% \bibliography{references}

{\small
\bibliographystyle{ieee_fullname}
\bibliography{references}

\begin{thebibliography}{10}\itemsep=-1pt

\bibitem{banner2019post}
Ron Banner, Yury Nahshan, and Daniel Soudry.
\newblock Post training 4-bit quantization of convolutional networks for
  rapid-deployment.
\newblock {\em Advances in Neural Information Processing Systems}, 32, 2019.

\bibitem{bengio2013estimating}
Yoshua Bengio, Nicholas L{\'e}onard, and Aaron Courville.
\newblock Estimating or propagating gradients through stochastic neurons for
  conditional computation.
\newblock {\em arXiv preprint arXiv:1308.3432}, 2013.

\bibitem{bhalgat2020lsq+}
Yash Bhalgat, Jinwon Lee, Markus Nagel, Tijmen Blankevoort, and Nojun Kwak.
\newblock Lsq+: Improving low-bit quantization through learnable offsets and
  better initialization.
\newblock In {\em Proceedings of the IEEE/CVF Conference on Computer Vision and
  Pattern Recognition Workshops}, pages 696--697, 2020.

\bibitem{boyd2011distributed}
Stephen Boyd, Neal Parikh, Eric Chu, Borja Peleato, Jonathan Eckstein, et~al.
\newblock Distributed optimization and statistical learning via the alternating
  direction method of multipliers.
\newblock {\em Foundations and Trends{\textregistered} in Machine learning},
  3(1):1--122, 2011.

\bibitem{jax2018github}
James Bradbury, Roy Frostig, Peter Hawkins, Matthew~James Johnson, Chris Leary,
  Dougal Maclaurin, George Necula, Adam Paszke, Jake Vander{P}las, Skye
  Wanderman-{M}ilne, and Qiao Zhang.
\newblock {JAX}: composable transformations of {P}ython+{N}um{P}y programs,
  2018.

\bibitem{chen2021vision}
Xiangning Chen, Cho-Jui Hsieh, and Boqing Gong.
\newblock When vision transformers outperform resnets without pre-training or
  strong data augmentations.
\newblock {\em arXiv preprint arXiv:2106.01548}, 2021.

\bibitem{chen2019eyeriss}
Yu-Hsin Chen, Tien-Ju Yang, Joel Emer, and Vivienne Sze.
\newblock Eyeriss v2: A flexible accelerator for emerging deep neural networks
  on mobile devices.
\newblock {\em IEEE Journal on Emerging and Selected Topics in Circuits and
  Systems}, 9(2):292--308, 2019.

\bibitem{choi2018pact}
Jungwook Choi, Zhuo Wang, Swagath Venkataramani, Pierce I-Jen Chuang,
  Vijayalakshmi Srinivasan, and Kailash Gopalakrishnan.
\newblock Pact: Parameterized clipping activation for quantized neural
  networks.
\newblock {\em arXiv preprint arXiv:1805.06085}, 2018.

\bibitem{dai2021vs}
Steve Dai, Rangha Venkatesan, Mark Ren, Brian Zimmer, William Dally, and Brucek
  Khailany.
\newblock Vs-quant: Per-vector scaled quantization for accurate low-precision
  neural network inference.
\newblock {\em Proceedings of Machine Learning and Systems}, 3:873--884, 2021.

\bibitem{deng2009imagenet}
Jia Deng, Wei Dong, Richard Socher, Li-Jia Li, Kai Li, and Li Fei-Fei.
\newblock Imagenet: A large-scale hierarchical image database.
\newblock In {\em 2009 IEEE conference on computer vision and pattern
  recognition}, pages 248--255. Ieee, 2009.

\bibitem{dong2020hawq}
Zhen Dong, Zhewei Yao, Daiyaan Arfeen, Amir Gholami, Michael~W Mahoney, and
  Kurt Keutzer.
\newblock Hawq-v2: Hessian aware trace-weighted quantization of neural
  networks.
\newblock {\em Advances in neural information processing systems},
  33:18518--18529, 2020.

\bibitem{elthakeb2018releq}
Ahmed~T Elthakeb, Prannoy Pilligundla, FatemehSadat Mireshghallah, Amir
  Yazdanbakhsh, and Hadi Esmaeilzadeh.
\newblock Releq: A reinforcement learning approach for deep quantization of
  neural networks.
\newblock {\em arXiv preprint arXiv:1811.01704}, 2018.

\bibitem{esser2019learned}
Steven~K Esser, Jeffrey~L McKinstry, Deepika Bablani, Rathinakumar Appuswamy,
  and Dharmendra~S Modha.
\newblock Learned step size quantization.
\newblock {\em arXiv preprint arXiv:1902.08153}, 2019.

\bibitem{garofalo2022darkside}
Angelo Garofalo, Yvan Tortorella, Matteo Perotti, Luca Valente, Alessandro
  Nadalini, Luca Benini, Davide Rossi, and Francesco Conti.
\newblock Darkside: A heterogeneous risc-v compute cluster for extreme-edge
  on-chip dnn inference and training.
\newblock {\em IEEE Open Journal of the Solid-State Circuits Society},
  2:231--243, 2022.

\bibitem{gholami2018squeezenext}
Amir Gholami, Kiseok Kwon, Bichen Wu, Zizheng Tai, Xiangyu Yue, Peter Jin,
  Sicheng Zhao, and Kurt Keutzer.
\newblock Squeezenext: Hardware-aware neural network design.
\newblock In {\em Proceedings of the IEEE Conference on Computer Vision and
  Pattern Recognition Workshops}, pages 1638--1647, 2018.

\bibitem{google_vizier}
Daniel Golovin, Benjamin Solnik, Subhodeep Moitra, Greg Kochanski, John Karro,
  and D. Sculley.
\newblock Google vizier: {A} service for black-box optimization.
\newblock In {\em Proceedings of the 23rd {ACM} {SIGKDD} International
  Conference on Knowledge Discovery and Data Mining, Halifax, NS, Canada,
  August 13 - 17, 2017}, pages 1487--1495. {ACM}, 2017.

\bibitem{gong2019differentiable}
Ruihao Gong, Xianglong Liu, Shenghu Jiang, Tianxiang Li, Peng Hu, Jiazhen Lin,
  Fengwei Yu, and Junjie Yan.
\newblock Differentiable soft quantization: Bridging full-precision and low-bit
  neural networks.
\newblock In {\em Proceedings of the IEEE/CVF International Conference on
  Computer Vision}, pages 4852--4861, 2019.

\bibitem{he2016deep}
Kaiming He, Xiangyu Zhang, Shaoqing Ren, and Jian Sun.
\newblock Deep residual learning for image recognition.
\newblock In {\em Proceedings of the IEEE conference on computer vision and
  pattern recognition}, pages 770--778, 2016.

\bibitem{flax2020github}
Jonathan Heek, Anselm Levskaya, Avital Oliver, Marvin Ritter, Bertrand
  Rondepierre, Andreas Steiner, and Marc van {Z}ee.
\newblock {F}lax: A neural network library and ecosystem for {JAX}, 2020.

\bibitem{optax2020github}
Matteo Hessel, David Budden, Fabio Viola, Mihaela Rosca, Eren Sezener, and Tom
  Hennigan.
\newblock Optax: composable gradient transformation and optimisation, in jax!,
  2020.

\bibitem{hinton2015distilling}
Geoffrey Hinton, Oriol Vinyals, Jeff Dean, et~al.
\newblock Distilling the knowledge in a neural network.
\newblock {\em arXiv preprint arXiv:1503.02531}, 2(7), 2015.

\bibitem{hooker2019compressed}
Sara Hooker, Aaron Courville, Gregory Clark, Yann Dauphin, and Andrea Frome.
\newblock What do compressed deep neural networks forget?
\newblock {\em arXiv preprint arXiv:1911.05248}, 2019.

\bibitem{ibrahim2021survey}
Ehab~M Ibrahim, Linyan Mei, and Marian Verhelst.
\newblock Survey and benchmarking of precision-scalable mac arrays for embedded
  dnn processing.
\newblock {\em arXiv preprint arXiv:2108.04773}, 2021.

\bibitem{jacob2018quantization}
Benoit Jacob, Skirmantas Kligys, Bo Chen, Menglong Zhu, Matthew Tang, Andrew
  Howard, Hartwig Adam, and Dmitry Kalenichenko.
\newblock Quantization and training of neural networks for efficient
  integer-arithmetic-only inference.
\newblock In {\em Proceedings of the IEEE conference on computer vision and
  pattern recognition}, pages 2704--2713, 2018.

\bibitem{jouppi2017datacenter}
Norman~P Jouppi, Cliff Young, Nishant Patil, David Patterson, Gaurav Agrawal,
  Raminder Bajwa, Sarah Bates, Suresh Bhatia, Nan Boden, Al Borchers, et~al.
\newblock In-datacenter performance analysis of a tensor processing unit.
\newblock In {\em Proceedings of the 44th annual international symposium on
  computer architecture}, pages 1--12, 2017.

\bibitem{jung2019learning}
Sangil Jung, Changyong Son, Seohyung Lee, Jinwoo Son, Jae-Joon Han, Youngjun
  Kwak, Sung~Ju Hwang, and Changkyu Choi.
\newblock Learning to quantize deep networks by optimizing quantization
  intervals with task loss.
\newblock In {\em Proceedings of the IEEE/CVF Conference on Computer Vision and
  Pattern Recognition}, pages 4350--4359, 2019.

\bibitem{kim2020position}
Jangho Kim, KiYoon Yoo, and Nojun Kwak.
\newblock Position-based scaled gradient for model quantization and pruning.
\newblock {\em Advances in Neural Information Processing Systems},
  33:20415--20426, 2020.

\bibitem{lee2021network}
Junghyup Lee, Dohyung Kim, and Bumsub Ham.
\newblock Network quantization with element-wise gradient scaling.
\newblock In {\em Proceedings of the IEEE/CVF Conference on Computer Vision and
  Pattern Recognition}, pages 6448--6457, 2021.

\bibitem{mo2022motuner}
Zewei Mo, Zejia Lin, Xianwei Zhang, and Yutong Lu.
\newblock motuner: a compiler-based auto-tuning approach for mixed-precision
  operators.
\newblock In {\em Proceedings of the 19th ACM International Conference on
  Computing Frontiers}, pages 94--102, 2022.

\bibitem{molendijk2022braintta}
Maarten Molendijk, Floran de Putter, Manil Gomony, Pekka
  J{\"a}{\"a}skel{\"a}inen, and Henk Corporaal.
\newblock Braintta: A 35 fj/op compiler programmable mixed-precision
  transport-triggered nn soc.
\newblock {\em arXiv preprint arXiv:2211.11331}, 2022.

\bibitem{nguyen2020quantization}
Hieu~Duy Nguyen, Anastasios Alexandridis, and Athanasios Mouchtaris.
\newblock Quantization aware training with absolute-cosine regularization for
  automatic speech recognition.
\newblock In {\em Interspeech}, pages 3366--3370, 2020.

\bibitem{park2020profit}
Eunhyeok Park and Sungjoo Yoo.
\newblock Profit: A novel training method for sub-4-bit mobilenet models.
\newblock In {\em European Conference on Computer Vision}, pages 430--446.
  Springer, 2020.

\bibitem{risso2023precision}
Matteo Risso, Alessio Burrello, Giuseppe~Maria Sarda, Luca Benini, Enrico
  Macii, Massimo Poncino, Marian Verhelst, and Daniele~Jahier Pagliari.
\newblock Precision-aware latency and energy balancing on multi-accelerator
  platforms for dnn inference.
\newblock {\em arXiv preprint arXiv:2306.05060}, 2023.

\bibitem{rutishauser2023free}
Georg Rutishauser, Francesco Conti, and Luca Benini.
\newblock Free bits: Latency optimization of mixed-precision quantized neural
  networks on the edge.
\newblock In {\em 2023 IEEE 5th International Conference on Artificial
  Intelligence Circuits and Systems (AICAS)}, pages 1--5. IEEE, 2023.

\bibitem{sandler2018mobilenetv2}
Mark Sandler, Andrew Howard, Menglong Zhu, Andrey Zhmoginov, and Liang-Chieh
  Chen.
\newblock Mobilenetv2: Inverted residuals and linear bottlenecks.
\newblock In {\em Proceedings of the IEEE conference on computer vision and
  pattern recognition}, pages 4510--4520, 2018.

\bibitem{shoeybi2019megatron}
Mohammad Shoeybi, Mostofa Patwary, Raul Puri, Patrick LeGresley, Jared Casper,
  and Bryan Catanzaro.
\newblock Megatron-{LM}: {T}raining multi-billion parameter language models
  using model parallelism.
\newblock {\em arXiv preprint arXiv:1909.08053}, 2019.

\bibitem{oss_vizier}
Xingyou Song, Sagi Perel, Chansoo Lee, Greg Kochanski, and Daniel Golovin.
\newblock Open source vizier: Distributed infrastructure and api for reliable
  and flexible black-box optimization.
\newblock In {\em Automated Machine Learning Conference, Systems Track
  (AutoML-Conf Systems)}, 2022.

\bibitem{tan2019efficientnet}
Mingxing Tan and Quoc Le.
\newblock Efficientnet: Rethinking model scaling for convolutional neural
  networks.
\newblock In {\em International conference on machine learning}, pages
  6105--6114. PMLR, 2019.

\bibitem{uhlich2019mixed}
Stefan Uhlich, Lukas Mauch, Fabien Cardinaux, Kazuki Yoshiyama, Javier~Alonso
  Garcia, Stephen Tiedemann, Thomas Kemp, and Akira Nakamura.
\newblock Mixed precision dnns: All you need is a good parametrization.
\newblock {\em arXiv preprint arXiv:1905.11452}, 2019.

\bibitem{wan2021edge}
Weier Wan, Rajkumar Kubendran, Clemens Schaefer, S~Burc Eryilmaz, Wenqiang
  Zhang, Dabin Wu, Stephen Deiss, Priyanka Raina, He Qian, Bin Gao, et~al.
\newblock Edge ai without compromise: Efficient, versatile and accurate
  neurocomputing in resistive random-access memory.
\newblock {\em arXiv preprint arXiv:2108.07879}, 2021.

\bibitem{wang2019haq}
Kuan Wang, Zhijian Liu, Yujun Lin, Ji Lin, and Song Han.
\newblock Haq: Hardware-aware automated quantization with mixed precision.
\newblock In {\em Proceedings of the IEEE/CVF Conference on Computer Vision and
  Pattern Recognition}, pages 8612--8620, 2019.

\bibitem{wu2019accelergy}
Yannan~Nellie Wu, Joel~S Emer, and Vivienne Sze.
\newblock Accelergy: An architecture-level energy estimation methodology for
  accelerator designs.
\newblock In {\em 2019 IEEE/ACM International Conference on Computer-Aided
  Design (ICCAD)}, pages 1--8. IEEE, 2019.

\bibitem{yao2021hawq}
Zhewei Yao, Zhen Dong, Zhangcheng Zheng, Amir Gholami, Jiali Yu, Eric Tan,
  Leyuan Wang, Qijing Huang, Yida Wang, Michael Mahoney, et~al.
\newblock Hawq-v3: Dyadic neural network quantization.
\newblock In {\em International Conference on Machine Learning}, pages
  11875--11886. PMLR, 2021.

\bibitem{yao2020pyhessian}
Zhewei Yao, Amir Gholami, Kurt Keutzer, and Michael~W Mahoney.
\newblock Pyhessian: Neural networks through the lens of the hessian.
\newblock In {\em 2020 IEEE international conference on big data (Big data)},
  pages 581--590. IEEE, 2020.

\bibitem{yazdanbakhsh2021evaluation}
Amir Yazdanbakhsh, Kiran Seshadri, Berkin Akin, James Laudon, and Ravi
  Narayanaswami.
\newblock An evaluation of edge tpu accelerators for convolutional neural
  networks.
\newblock {\em arXiv preprint arXiv:2102.10423}, 2021.

\bibitem{zhai2021scaling}
Xiaohua Zhai, Alexander Kolesnikov, Neil Houlsby, and Lucas Beyer.
\newblock Scaling vision transformers.
\newblock {\em arXiv preprint arXiv:2106.04560}, 2021.

\bibitem{zhang2022pokebnn}
Yichi Zhang, Zhiru Zhang, and Lukasz Lew.
\newblock Pokebnn: A binary pursuit of lightweight accuracy.
\newblock In {\em Proceedings of the IEEE/CVF Conference on Computer Vision and
  Pattern Recognition}, pages 12475--12485, 2022.

\end{thebibliography}
}

%%%%%%%%%%%%%%%%%%%%%%%%%%%%%%%%%%%%%%%%%%%%%%%%%%%%%%%%%%%%

%%%%%%%%%%%%%%%%%%%%%%%%%%%%%%%%%%%%%%%%%%%%%%%%%%%%%%%%%%%%
% everything below here is not due on Thursday.
\newpage

\appendix
\section{Supplementary Materials}\label{calib-text}
We introduced our quantization method in Sections~\ref{sec:back} and~\ref{sec:meth} of the main paper. Here, we will cover various sweeps, sensitivity analysis, and ablations that we have performed to more comprehensively study our quantization scheme. 
\subsection{Calibration Methods}

Model calibration refers to the process of determining a good step size ($d$) and dynamic range ($q_{+}$) for the quantizer (see eq.~\eqref{eq:1} for how we define these terms). Previous studies have shown that not only does the calibration method impact the final quantization outcome, the same method might not be optimal across different bitwidths~\cite{dai2021vs}. Additionally, weights and activations have shown different sensitivities to quantization and dynamic range in computation, notably recent mixed precision work~\cite{dong2020hawq} seem to employ higher precision on activations. Consequently, we study how different calibration schemes might impact quantization performance through a coarse grid-search studying the quantization performance on a 3-bit homogeneous EfficientNet-Lite0 followed by quantization aware training for 50 epochs. Our results are summarized in Table~\ref{sample-table}, which show: (i) calibration method impact for weights vs. activations and (ii) calibration method impact at different bit precisions. The first phase of our training implements a homogeneously quantized model, with the dynamic range and step size being determined through the data of the first batch. We implement the following baselines to determine the clipping value ($q_\text{max}$): (i) the maximum value in the data, (ii) $2\times$ the mean of the data, (iii) $\max(\mu+3\sigma,|\mu-3\sigma|)$ (Gaussian in table~\ref{sample-table}), and (iv) calibrating to the $X^\mathrm{th}$ percentile of the absolute value of the data over a range~\cite{dai2021vs}. The choice of calibration scheme results in a $2.05\%$ improvement over alternatives. The effect of different calibration schemes gets less pronounced at the higher precision (4 bits vs. 3 bits --- see Table~\ref{appendix:calib_4b} for 4 bits).

%offer the worst calibration scheme (excluding combinations which lead to accuracy $\leq 1.0\%$). The best calibration scheme is using the Gaussian method on weights and the $99.9^\mathrm{th}$ of the activations, supporting the theory that activations and weights should be treated differently when it comes to quantization.

\begin{table*}
  \caption{Effect of different calibration methods on homogeneous bit-width training (3 bits) for a Efficient-Lite0 on the ImageNet dataset.}
  \label{sample-table}
  \centering
\begin{tabular}{lrrrrrrr}
\toprule
W/A         & \multicolumn{1}{l}{Max}         & \multicolumn{1}{l}{$2\times$ Mean} & \multicolumn{1}{l}{Gaussian}    & \multicolumn{1}{l}{P99.9}       & \multicolumn{1}{l}{P99.99}      & \multicolumn{1}{l}{P99.999}     & \multicolumn{1}{l}{P99.9999}    \\
\midrule
Max         & 63.85\% & 63.78\% & 63.38\% & 63.24\% & 63.59\% & 63.73\% & 63.54\% \\
$2\times$ Mean & 64.57\% & 0.10\%  & 63.02\% & 64.51\% & 64.55\% & 64.72\% & 64.22\% \\
Gaussian    & 64.72\% & 64.73\% & 64.90\% & \textbf{65.07\%} & 64.75\% & 64.67\% & 64.79\% \\
P99.9       & 64.67\% & 64.87\% & 64.64\% & 64.71\% & 64.64\% & 64.86\% & 64.88\% \\
P99.99      & 64.56\% & 64.81\% & 64.44\% & 64.27\% & 64.45\% & 64.33\% & 64.51\% \\
P99.999     & 64.29\% & 64.59\% & 64.50\% & 64.38\% & 63.96\% & 64.03\% & 64.25\% \\
P99.9999    & 64.14\% & 64.23\% & 63.78\% & 63.73\% & 63.95\% & 64.34\% & 64.26\% \\
\bottomrule
\end{tabular}
\label{appendix:calib}
\end{table*}

\begin{table*}
  \caption{Effect of different calibration methods on homogeneous bit-width training (4 bits) for a Efficient-Lite0 on the ImageNet dataset.}
  \label{appendix:calib_4b}
  \centering
\begin{tabular}{lrrrrrrr}
\toprule
W/A         & \multicolumn{1}{l}{Max}         & \multicolumn{1}{l}{$2\times$ Mean} & \multicolumn{1}{l}{Gaussian}    & \multicolumn{1}{l}{P99.9}       & \multicolumn{1}{l}{P99.99}      & \multicolumn{1}{l}{P99.999}     & \multicolumn{1}{l}{P99.9999}    \\
\midrule
Max         & 72.08\% & 71.96\% & 71.89\% & 72.00\% & 71.87\% & 71.95\% & 72.08\% \\
$2\times$ Mean & 72.25\% & 0.73\%  & 72.09\% & 72.22\% & 72.15\% & 72.29\% & 72.30\% \\
Gaussian    & 72.24\% & 72.27\% & 72.15\% & 72.26\% & 72.29\% & 72.29\% & 72.38\% \\
P99.9       & 72.29\% & 72.44\% & 72.20\% & 72.31\% & 72.17\% & 72.12\% & 72.44\% \\
P99.99      & 72.10\% & 72.10\% & 72.10\% & 72.32\% & 72.21\% & 72.27\% & 72.27\% \\
P99.999     & 72.21\% & 72.11\% & 72.03\% & 71.95\% & 72.16\% & 72.25\% & 72.19\% \\
P99.9999    & 72.09\% & 72.01\% & 72.10\% & 72.05\% & 71.97\% & 72.25\% & 72.02\% \\
\bottomrule
\end{tabular}
\end{table*}

\subsection{Comparison of Gradient Scaling Methods}\label{asec:grad_scale}

\begin{figure*}
  \centering
  \includegraphics[width=.75\textwidth]{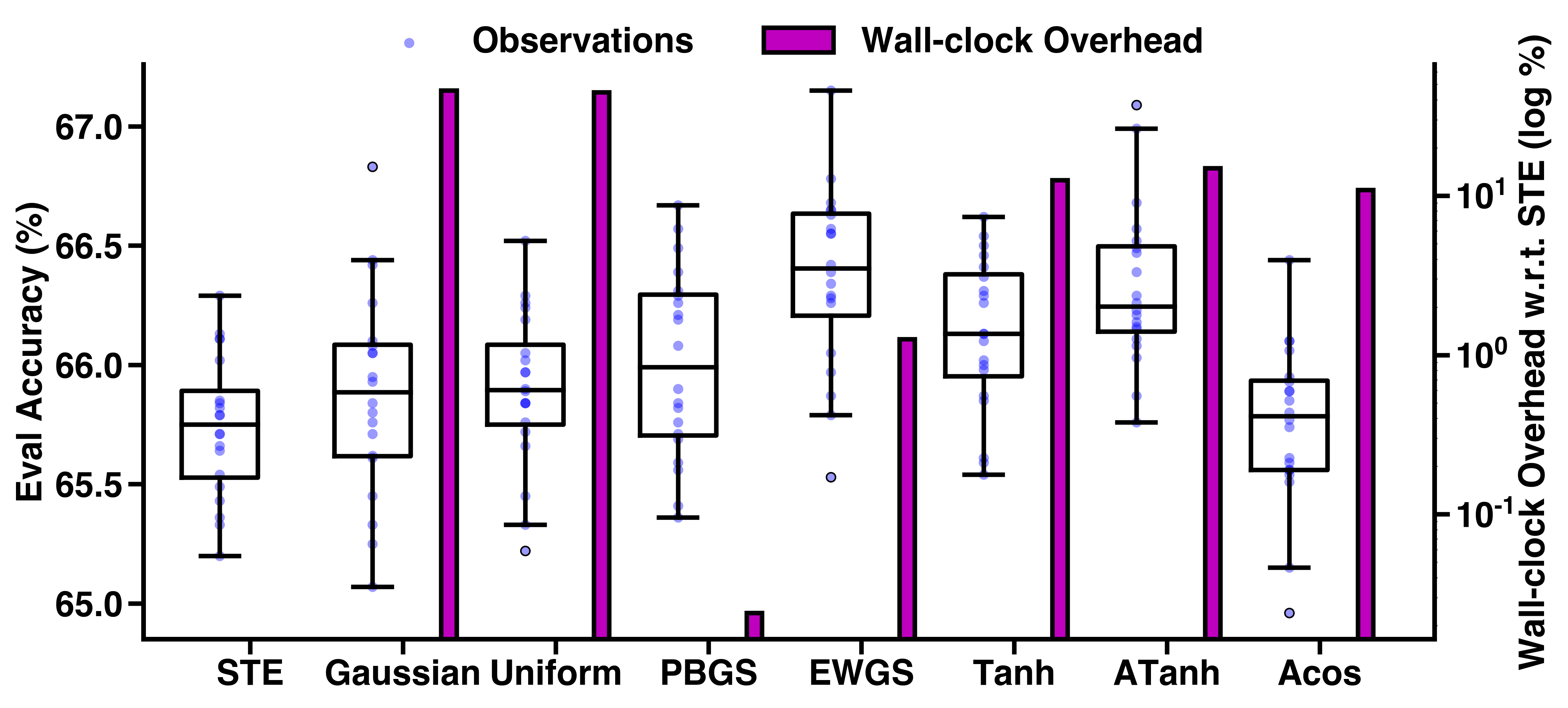}
  %\fbox{\rule[-.5cm]{0cm}{4cm} \rule[-.5cm]{4cm}{0cm}}
  \caption{Comparison of gradient scaling methods on a EfficientNet-Lite0 with weights and activation both quantized to 3 bits with a gradient scale factor of 5e-3. STE~(\cite{bengio2013estimating}) stands for the straight-through-estimator which does not modify the gradient meanwhile Gaussian and Uniform add random noise to the gradient in the backward pass. PBGS~(\cite{kim2020position}), EWGS~(\cite{lee2021network}) and Acos~(\cite{nguyen2020quantization}) scale the gradient based on the distance from the quantization point. We added two additional position based scaling methods Tanh and InvTanh. We also display the wall-clock time overhead of gradient scaling method in log percent compared to STE.} 
  \label{fig:uniform}
\end{figure*}

In the main text we showed results for gradient scaling when using different gradient scaling methods for weights and activations, in Figure~\ref{fig:uniform} we show the baseline results when the scaling methods are the same for weights and activations. The mean for STE and EWGS (best method in this experiment) also serve as orientations in Figure~\ref{fig:98}. We further show the effect of adding noise to the gradient instead of meaningful gradient scaling (Gaussian and uniform noise), both methods boost the accuracy beyond STE accuracy on average. Further, we show the computational overhead measured in wall-clock time compare to the STE method, which does not perform any gradient modification. The methods which employ random noise are the most computationally expensive compared to the STE. Interestingly, PBGS only incurs minimal overhead while the other methods show a substantial increase in compute time. We conjecture that this can be primarily be attributed to the trigonometric operation in the backward pass.  

\begin{figure*}
  \centering
  \includegraphics[width=.75\linewidth]{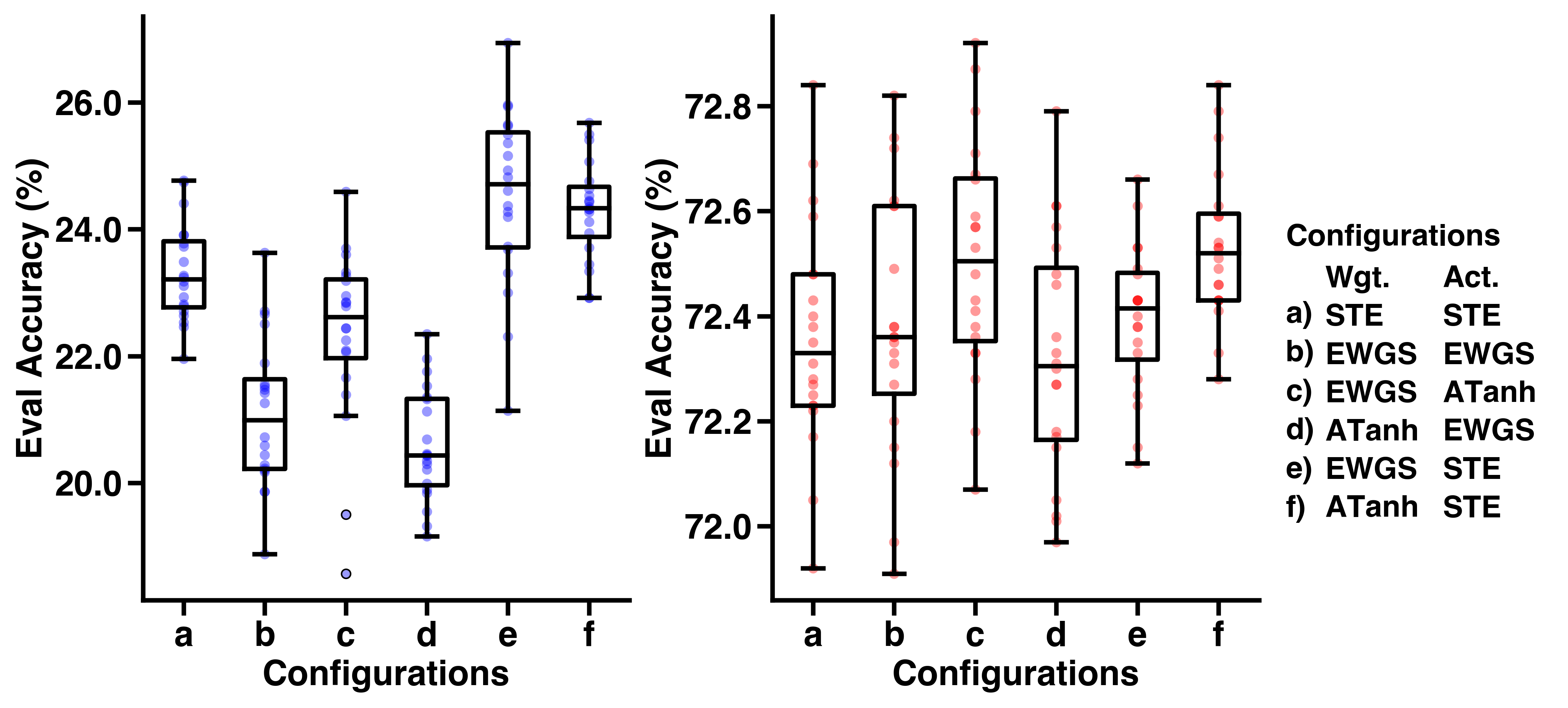}
\caption{Comparison of selected gradient scaling methods on a homogeneously quantized 2 and 4 bit EfficientNet-Lite0 network. Accuracy numbers drop drastically for 2 bit networks and gradient scaling methods on activations show lower performance compared to the straight-through estimator method.}
\label{fig:gradscale-sweeps}
\end{figure*}

We also analyzed the behaviour of several hand-picked scaling methods on lower and higher bit widths comparable to what we have done in Figure~\ref{fig:98}. The results can be seen in Figure~\ref{fig:gradscale-sweeps}, showing that at 2 bits no method delivered better than $30\%$ accuracy. The differences between these methods are less apparent ($\leq0.4\%$) at 4 bits. 

\subsection{Effect of Granularity and Gradient Scaling}\label{asec:morers}

We demonstrate the effectiveness of our proposed quantization (see config. j in Figure~\ref{fig:gradscale-sweeps}), through ablation trials where we remove gradient scaling, fine-grained quantization, and their combination (base in~\ref{asec:morers}). Across different models and target budgets, we see that our proposed fine-grained scheme consistently occupies the Pareto, although not all trials perform as well (~\ref{asec:morers}).

\begin{figure*}
  \centering
  \includegraphics[width=.75\textwidth]{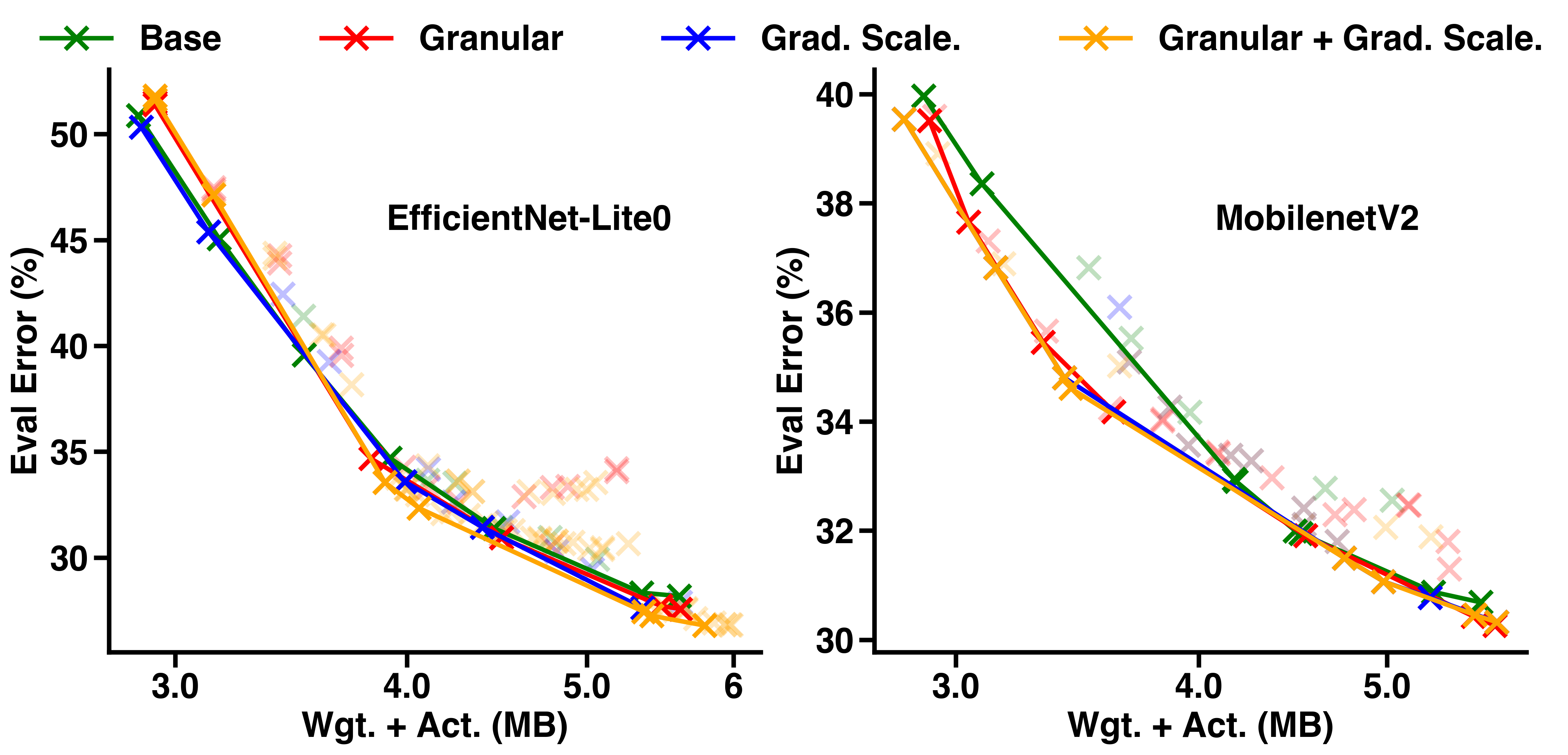} % \label{fig:enet_mbnet}
  %\fbox{\rule[-.5cm]{0cm}{4cm} \rule[-.5cm]{4cm}{0cm}}
  \caption{EfficientNet-Lite0 and MobileNetV2 mixed precision results showing all obtained data points, even those which are not Pareto optimal.} 
  \label{fig:sweeps_enet_mbnet}
\end{figure*}

\begin{figure*}
  \centering
  \includegraphics[width=.75\textwidth]{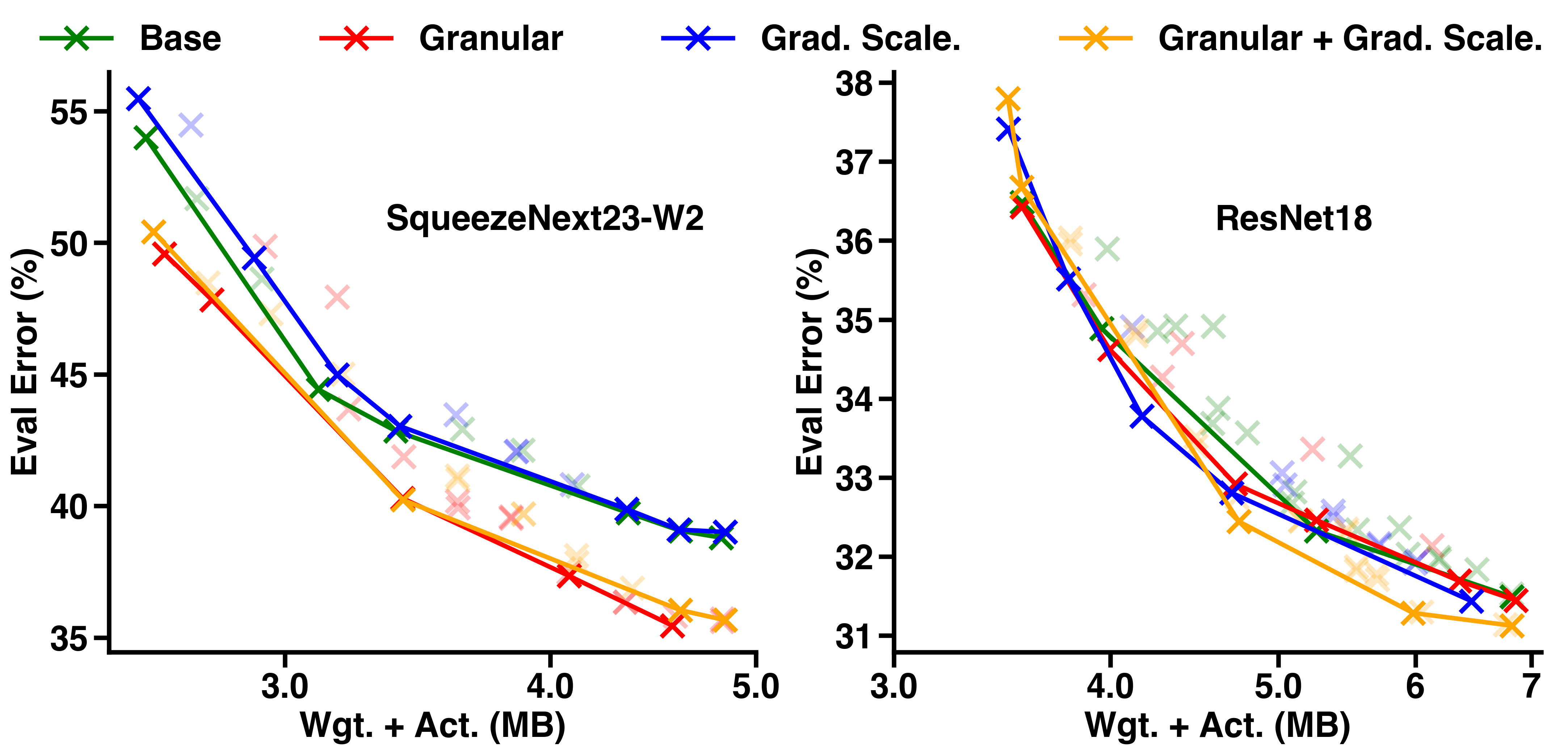} % \label{fig:sq_resnet}
  %\fbox{\rule[-.5cm]{0cm}{4cm} \rule[-.5cm]{4cm}{0cm}}
  \caption{SqueezeNext23-W2 and ResNet18 mixed precision results showing all obtained data points, even those which are not Pareto optimal.} 
  \label{fig:sweeps_sqnxt_resnet}
\end{figure*}

\subsection{Assumption for Comparison to Other Work}

To facilitate our comparisons in (Figure~\ref{fig:overview}), we computed the memory footprint for weights and activations for different networks, assuming floating point (bfloat16) parameters for batch norm. If the references stated that they did not quantize the first and last layer (~\cite{choi2018pact, esser2019learned, bhalgat2020lsq+, lee2021network, jung2019learning}), we assume they can use bfloat16 datatypes without any loss of accuracy. Table~\ref{tab:asum:} states our assumptions about networks sizes in terms of how many parameters models have in total, how many of those parameters are for matrix-vector multiplications, how many parameters are batch norm (BN) parameters and how many parameters belong to the first and last layers.

\begin{table*}
  \caption{Numbers used to compute effective network sizes for other works.}
  \label{tab:asum:}
  \centering
  \begin{tabular}{lrrrrr}
    \toprule
    %\multicolumn{2}{c}{Part}                   \\
    & ResNet18 & MobileNetV2 & SqNxt23-W2 & ENet-B0 & ENet-Lite0 \\
    %\cmidrule(r){1-2}
    %Name     & Description     & Size ($\mu$m) \\
    \midrule
    \# Total Wgt. & 11,689,512 & 3,504,872 & 20,670,016 & 5,288,548 & 4,652,008  \\
    \# MVM Wgt. & 11,679,912 & 3,470,760 & 20,485,184 & 5,246,532 & 4,609,992 \\
    \# First Layer Wgt. & 9,408 & 864 & 18,816 & 864 & 864 \\
    \# Last Layer Wgt. & 513,000 & 1,281,000 & 256,000& 1,281,000 & 1,281,000 \\
    \# BN Wgt. & 9,600 & 34,112 & 184,832 & 42,016 & 42,016 \\
    \midrule
    \# Total Sum Act. & 2,032,640 & 6,678,112 & 3,962,208 & 8,982,784 & 6,676,256 \\
    \# Last Layer Act. & 512 & 1,280 & 256 & 1,280 & 1,280 \\
    \bottomrule
  \end{tabular}
\end{table*}

\subsection{Latency Considerations}\label{app-lat}

To highlight the possible effects on latency we simulated latency numbers from synthesized single MAC units on a commercially available advanced node. Realistic latency numbers require a comprehensive co-design between the hardware and accelerator, including parameters like technology node, hardware, compiler configurations etc. Note that these are weak approximations and a more comprehensive evaluation will require full architectural simulation. Consider that data movement between memory hierarchies and computing units has a major impact on latency. Minimizing this through quantization will impact the end-to-end latency. For example, consider a hypothetical accelerator with a bandwidth of 128 bits/cycle to transfer data between the global buffer and local register. A convolution layer of size 3x3x128 will require 4x the number of cycles to transfer 16bits (144 cycles) vs. 4 bits (36 cycles). While some amount of this can be hidden through communication-computation overlap, quantization will still reduce the number of cycles for which computing might stall while waiting for additional data. In our simulations each MAC unit can have different bit-width inputs. To obtain a full model latency we added up different latencies of single MAC units per layer given our trained bit-width. Table~\ref{tab:latency} shows the results for our frontier results (Efficient-Lite0 and MobileNetV2). The latency results are relative to a 16 bit model.

\begin{table*}
  \caption{Relative simulated latency numbers based on single mixed precision MAC  units.}
  \label{tab:latency}
  \centering
\begin{tabular}{lrrrrrrrrr}
\toprule
\multicolumn{10}{c}{EfficientNet-Lite0}                                                                 \\
\midrule
Size (MB)     & 22.66  & - & 3.01  & 3.23  & 3.98  & 4.14  & 5.45  & 5.50  & 5.87  \\
Accuracy (\%)      & 75.53 & - & 48.37 & 52.87 & 66.46 & 67.66 & 72.56 & 72.75 & 73.21 \\
%Normalized Latency & 250.0 & - &  77.60 & 78.44 & 84.22 & 84.66 & 97.27 & 96.86 & 105.39 \\
Norm. Latency (\%) & 100.00&	-&	31.04&	31.38&	33.69&	33.86&	38.91&	38.74&	42.16 \\
% ACE  &   &   &   &   &   &   &   &   \\
% \bottomrule
% \vspace{.45em}
% \end{tabular}

% \begin{tabular}{lrrrrrrrrr}
% \toprule
\midrule
\multicolumn{10}{c}{MobileNetV2}                                                                        \\
\midrule
Size (MB)     & 20.25 & 2.89  & 3.21  & 3.48  & 3.51  & 4.82  & 5.05  & 5.62  & 5.76  \\
Accuracy (\%)  & 71.46 & 60.72 & 63.18 & 65.20 & 65.39 & 68.50 & 68.93 & 69.54 & 69.68 \\
%Norm. Latency & 265.0 & 88.65  & 87.62  & 91.84  & 94.71  & 99.05  & 101.09 & 116.45 &  110.32 \\
Norm. Latency (\%) & 100.00 &	33.45&	33.06&	34.66&	35.74&	37.38&	38.15&	43.94&	41.63 \\
% ACE  &  &  &   &   &  &  &  &  &  \\
\bottomrule
\end{tabular}
\end{table*}

\subsection{ADMM for Heterogenous Quantization}\label{asec:admm}

% \begin{equation} \label{eq:2}
% \begin{aligned}
%    L = & CE(x, y) + \beta \max \left( \left( \sum^L_{l=1} \sum^C_{c=1} b^w_{lc} \cdot s_{lc}^w \right)  - t^w ,0 \right)^2\\
%    & + \beta \max \left( \left( \sum^L_{l=1} b^a_l \cdot s_l^a\right) - t^a, 0\right)^2.
% \end{aligned}
% \end{equation}

To apply ADMM to quantization outlined in \sj{()}, we reformulate the problem as:

\begin{equation} \label{eq:admm}
\begin{aligned}
 \min_{x_1,x_2} & \quad CE(x_1, y) \\
 & + \beta \max \left( \left( \sum^L_{l=1} \sum^C_{c=1} b^{w}_{lc, x_2} \cdot s_{lc}^w \right)  - t^w ,0 \right)^2 \\
 & + \beta \max \left( \left( \sum^L_{l=1} b^a_{l,x_2} \cdot s_l^a\right) - t^a, 0\right)^2 \\
& \textrm{s.t.} \quad x_1 - x_2 = 0,
\end{aligned}
\end{equation}
where the notation is the same as in eq.~\eqref{eq:2}. However, due to ADMM's alternating phase formulation, we now have two sets of model weights $x_1$ and $x_2$. The first set of model weights are used only for the accuracy optimization (see CE loss) and the second part only for size penalties. Given the constraint, we can form an augmented Lagrangian:

\begin{equation} \label{eq:lagrange_admm}
\begin{aligned}
F(x_1,x_2,y) =  & \quad CE(x_1, y) \\
 & + \beta \max \left( \left( \sum^L_{l=1} \sum^C_{c=1} b^w_{lc, x_2} \cdot s_{lc}^w \right)  - t^w ,0 \right)^2 \\
 & + \beta \max \left( \left( \sum^L_{l=1} b^a_{l, x_2} \cdot s_l^a\right) - t^a, 0\right)^2 \\
& + y^T (x_1-x_2) + \frac{\rho}{2}\lVert x_1-x_2\rVert^2_2.
\end{aligned}
\end{equation}
% \begin{equation} \label{eq:lagrange_admm}
%  F(x,z,y) = f(x) + g(z) + y^T h(x,z) + \frac{\rho}{2}\lVert h(x,z)\rVert^2_2
% \end{equation}

To which we can apply the ADMM method consisting of the following steps:

\begin{enumerate}
    \item $ \bar{x_1} = \mathrm{arg}\min_{x_1} F(x_1,x_2,y) $ minimize $x_1$ while $x_2$ and $y$ kept constant.
    \item $ \bar{x_2} = \mathrm{arg}\min_{x_2} F(\bar{x_1},x_2,y) $ minimize $x_2$ while $y$ and $\bar{x_1}$ kept constant.
    \item $\bar{y} = y + \rho h(\bar{x_1}, \bar{x_2})$ update multiplier.
\end{enumerate}

ADMM performance is sensitive to the learning rate in steps 1 and 2, and the hyperparameter $\rho$. To overcome this, we conducted an extensive hyperparameter search using a BBO algorithm~\cite{oss_vizier, google_vizier}. Based on this search. $\rho$ was set to $18.1179$ and the learning rate was set to $8.95062e^{-6}$ for an EfficientNet-Lite0 with a budget equivalent to a 4-bit model.

% 1. $ \bar{x} = \mathrm{arg}\min_{x} F(x,y,z) $ minimize $x$ while $y$ and $z$ kept constant.
% 2. $ \bar{z} = \mathrm{arg}\min_{z} F(\bar{x},y,z) $ minimize $z$ while $y$ and $\bar{x}$ kept constant.
% 3. $\bar{y} = y + \rho h(\bar{x}, \bar{z})$ update multiplier.

% ADMM for Heterogenous Quantization

% $$ \min_{x,z} \quad \mathrm{Cross\; Entropy}(x) + \mathrm{Size\; Loss}(z) \\
% \textrm{s.t.} \quad x - z = 0$$

\subsection{Limitations}
The main limitation of our work is that to achieve a specific model size we use standard gradient descent optimization with penalties on the loss function which does not guarantee that the model size constraint will be fulfilled. The size constraints are part of the loss function and with a sufficiently large penalty factor ($\beta$) the chance of fulfilling the constraint can substantially increase but simultaneously performance might be sacrificed. Furthermore, our method does not guarantee a solution on the efficient frontier. See~\ref{asec:morers} in the supplementary materials where we summarize the sensitivity of different elements of our technique across multiple training runs. Typical solutions are within 1\% or on the Pareto frontier however some memory constraints prove  especially hard to quantize efficiently on some networks, e.g. EfficientLite-0 around 5 MB yields sub-optimal solutions at two different accuracy levels both more than 1\% away from the Pareto frontier. Due to convergence limitations, we did not examine techniques for binarization and terenerization which could further improve the memory and energy footprint of these models. 

Like related work, we do not quantize batch normalization parameters which contribute significantly to the model size post-quantization. Techniques to ``fold'' these parameters  into the convolution or affine layers have been proposed that could be leveraged to minimize this~\cite{jacob2018quantization}.

\subsection{Societal Impact}

Our work describes a general technique to train mixed precision neural networks and demonstrates its performance with image classification on the ImageNet dataset. In principle, our techniques can be applied to other domains as well. Our quantization method trades off accuracy for model size. We have, however, not analyzed what the model forgets when its size constraints are tightened. Other research has recently investigated this issue for model compression~\cite{hooker2019compressed}. A similar analysis for mixed precision quantization remains open. We foresee the primary impact on society being a reduction in the cost of deploying machine learning systems on edge devices, reducing energy consumption and carbon footprint. However, widespread deployment of ML on such devices could see negative uses, e.g., surveillance. Additional negative impacts might arise due to the need for mixed-precision accelerators which might increase the barrier to entry for deploying such models.

\subsection{Effect of Different Training Phases}

Table~\ref{tab:corr_bbo} shows Pearson's correlation coefficient of the final accuracy and parameters from the BBO (note we only included runs that actually met our size constraints). The correlation coefficients give insights into the importance of the training phases and settings. For example, the activation penalty has a high correlation coefficient, hinting at its importance in training. For further phase ablation studies, we would like to note that ablating the phase itself (i.e., removing the phase itself) might not be particularly meaningful. In our initial studies, removing phase 1 (homogeneous network training) resulted in near-random accuracy. Removing phase 2 removes the bit-width learning mechanism, thereby preventing adaptation to size targets. Removing phase 3 prevents performance recovery from finetuning, which generally yielded a 2-3\% improvement in final accuracy.

\begin{table}
  \caption{Correlation coefficients of BBO search space parameters to accuracy.}
  \label{tab:corr_bbo}
  \centering
\begin{tabular}{lc}
\toprule
Parameter & Corr. Coef.\\
\midrule
Activation Penalty & -0.0756 \\
Weight Penalty & -0.0567 \\ 
Penalty Ramp-up Length & -0.0491 \\
Update Frequency & +0.0408 \\
\bottomrule
\end{tabular}
\end{table}

\end{document}